\documentclass[journal]{IEEEtran}

\usepackage{ntheorem}
\theoremheaderfont{\rmfamily\bfseries\upshape}
\theorembodyfont{\rmfamily\mdseries\itshape}
\theoremseparator{.}

\usepackage{bbm}
\usepackage{upgreek}
\usepackage{amsmath}
\usepackage{xcolor}
\usepackage{graphicx}
\usepackage{booktabs}
\usepackage{amssymb}
\usepackage{textcomp}
\usepackage{amsmath,bm}
\usepackage{makecell}
\usepackage{multirow}
\usepackage{caption}
\usepackage{float}
\usepackage{colortbl}
\usepackage{mathrsfs}
\usepackage{changepage}
\usepackage[colorlinks]{hyperref}
\usepackage{color}
\usepackage{soul}
\usepackage{tcolorbox}
\usepackage{subfigure}
\usepackage{tikz}
\usepackage{lipsum}
\usepackage[mathlines,switch]{lineno}
\usetikzlibrary{calc,tikzmark}
\usepackage[numbers,sort&compress]{natbib}
\makeatletter  
\newif\if@restonecol  
\makeatother

\usepackage[linesnumbered,ruled,vlined]{algorithm2e}%[ruled,vlined]{  
\usepackage{algpseudocode}
\usepackage{amsmath}  
\definecolor{myboxcolor}{rgb}{1, 1, 0.96}
\definecolor{myred}{rgb}{1, 0, 0}

\usepackage{tikz,xcolor,hyperref}% Make Orcid icon
\definecolor{lime}{HTML}{A6CE39}
\DeclareRobustCommand{\orcidicon}{%
    \begin{tikzpicture}
    \draw[lime, fill=lime] (0,0) 
    circle [radius=0.16] 
    node[white] {{\fontfamily{qag}\selectfont \tiny ID}};    \draw[white, fill=white] (-0.0625,0.095)
    circle [radius=0.007];    \end{tikzpicture}
    \hspace{-2mm}}
\foreach \x in {A, ..., Z}{%
    \expandafter\xdef\csname orcid\x\endcsname{\noexpand\href{https://orcid.org/\csname orcidauthor\x\endcsname}{\noexpand\orcidicon}}
    }
\hypersetup{hidelinks}
\ifCLASSINFOpdf

\else

\fi

\begin{document}
%\linenumbers
\IEEEoverridecommandlockouts
\IEEEpubid{\begin{minipage}[t]{\textwidth}\ \\[0.6pt]
        \centering\footnotesize{\begin{tcolorbox}[left = 0.5mm, right = 0.5mm, top = 0.5mm, bottom = 0.5mm]\copyright This work has been submitted to the IEEE for possible publication. Copyright may be transferred without notice, after which this version may no longer be accessible.\end{tcolorbox}}
\end{minipage}}

\title{Multi-Task Surrogate-Assisted Search with Bayesian Competitive Knowledge Transfer for Expensive Optimization}

\author{Yi Lu,
        Xiaoming Xue\orcidA{}, \IEEEmembership{~Member,~IEEE},
        Kai Zhang\orcidB{}, \IEEEmembership{~Member,~IEEE},
        Liming Zhang,\\
        Guodong Chen\orcidC{},
        Chenming Cao\orcidD{},
        Piyang Liu,
        and Kay Chen Tan\orcidE{},~\IEEEmembership{~Fellow,~IEEE}
\thanks{Yi Lu, Kai Zhang and Piyang Liu are with the Civil Engineering School, Qingdao University of Technology, Qingdao 266520, China (e-mail: lyqut@outlook.com; zhangkai@qut.edu.cn; piyang.liu@qut.edu.cn).}% <-this % stops a space
\thanks{Xiaoming Xue, Chenming Cao and Kay Chen Tan are with the Department of Data Science and Artificial Intelligence, The Hong Kong Polytechnic University, Hong Kong SAR, China (e-mail: xiaoming.xue@polyu.edu.hk; ccm831143@outlook.com; kctan@polyu.edu.hk).}% <- this % stops a space
\thanks{Liming Zhang is with the School of Petroleum Engineering, China University of Petroleum (East China), Qingdao 266580, China (e-mail: zhangliming@upc.edu.cn).}
\thanks{Guodong Chen is with the Department of Earth Sciences, The University of Hong Kong, Hong Kong SAR, China (e-mail: u3008598@connect.hku.hk).}% <- this % stops a space
}% <-this % stops a space

%\textcolor{blue}{
% make the title area
\maketitle

\begin{abstract}
Expensive optimization problems (EOPs) present significant challenges for traditional evolutionary optimization due to their limited evaluation calls.
Although surrogate-assisted search (SAS) has become a popular paradigm for addressing EOPs, it still suffers from the cold-start issue.
In response to this challenge, knowledge transfer has been gaining popularity for its ability to leverage search experience from potentially related instances, ultimately facilitating head-start optimization for more efficient decision-making.
However, the curse of negative transfer persists when applying knowledge transfer to EOPs, primarily due to the inherent limitations of existing methods in assessing knowledge transferability.
On the one hand, a priori transferability assessment criteria are intrinsically inaccurate due to their imprecise understandings.
On the other hand, a posteriori methods often necessitate sufficient observations to make correct inferences, rendering them inefficient when applied to EOPs.
Considering the above, this paper introduces a Bayesian competitive knowledge transfer (BCKT) method developed to improve multi-task SAS (MSAS) when addressing multiple EOPs simultaneously.
Specifically, the transferability of knowledge is estimated from a Bayesian perspective that accommodates both prior beliefs and empirical evidence, enabling accurate competition between inner-task and inter-task solutions, ultimately leading to the adaptive use of promising solutions while effectively suppressing inferior ones.
The effectiveness of our method in boosting various SAS algorithms for both multi-task and many-task problems is empirically validated, complemented by comparative studies that demonstrate its superiority over peer algorithms and its applicability to real-world scenarios.
The source code of our method is available at \textcolor{magenta}{\url{https://github.com/XmingHsueh/MSAS-BCKT}}.
\end{abstract}

\begin{IEEEkeywords}
evolutionary transfer optimization, surrogate-assisted search, Bayesian inference, when to transfer.
\end{IEEEkeywords}

\IEEEpeerreviewmaketitle

%################################################################################
%######################################divider######################################
%################################################################################

\section{Introduction}
\label{section:intro}

Expensive optimization problems (EOPs) pose challenges for most optimizers, as evaluating the objective functions or constraints always requires substantial resources or time, thereby restricting the total number of function evaluations (FEs) available during the optimization process~\cite{ong2003evolutionary,jin2018data}.
Such problems are prevalent across various domains, including hyperparameter optimization in computer science~\cite{falkner2018bohb}, reservoir production optimization in petroleum engineering~\cite{dai2025adaptive}, conformational search in computational chemistry~\cite{watts2010confgen}, and processing parameter optimization in material design~\cite{khatamsaz2023physics}, among others.
Although evolutionary algorithms (EAs) have proven effective in solving complex optimization tasks due to their global search capabilities and adaptability, this effectiveness often comes at the expense of a large number of evaluation calls consumed by a series of solution sets (i.e., populations), making EAs computationally unaffordable for many EOPs.

Over the decades, a variety of techniques haven been developed to improve the search efficiency of EAs, including surrogate modeling~\cite{lim2009generalizing}, knowledge transfer~\cite{gupta2015multifactorial}, divide and conquer~\cite{yang2017turning}, learning-aided evolution~\cite{liu2022learning}, operator enhancement~\cite{zychowski2018addressing}, and parallel computing~\cite{alba2002parallelism}, among others.
Significantly, surrogate-assisted search (SAS)~\cite{jin2011surrogate} and evolutionary transfer optimization (ETO)~\cite{tan2021evolutionary} have been gaining popularity in research due to their extensive successes reported in various domains.
In particular, SAS utilizes computationally efficient surrogate models to replace the original EOPs, enabling effective optimization through substantial cheap evaluations~\cite{si2023linear}.
Commonly used surrogate models include polynomial response surface~\cite{box1978statistics}, Gaussian process regression~\cite{williams2006gaussian}, and radial basis function~\cite{hardy1971multiquadric}.
By contrast, ETO aims to integrate EAs with knowledge learning and transfer across related domains for better optimization performance, which mainly consists of three conceptual realizations~\cite{gupta2017insights}, namely sequential transfer~\cite{feng2017autoencoding,cao2024global}, multitasking~\cite{gupta2015multifactorial,wu2024evolution}, and multiform optimization~\cite{chen2022scaling,feng2024review}.
In recent years, there has been an emerging trend of seamless fusions of SAS and ETO~\cite{min2017multiproblem,yang2019offline,huang2021surrogate,liu2024extremo,li2024fast,tan2024surrogate,shen2024surrogate,xue2024surrogate}, which are considered promising solutions for addressing the cold start issue in EOPs.
This trend drives this study to move toward multi-task SAS (MSAS) that can achieve adaptive knowledge transfer among multiple EOPs solved concurrently.

When integrating knowledge learning and transfer capabilities into SAS to create MSAS, three fundamental issues must be addressed~\cite{xue2023solution}: what to transfer, how to transfer, and when to transfer.
In regard to what to transfer, one needs to determine the form of knowledge to be transferred, such as solutions~\cite{shen2024surrogate}, parameters~\cite{wang2024distilling}, etc.
Moreover, when multiple knowledge candidates are available, the most promising one should be identified for transfer with certain source selection strategies~\cite{chen2020adaptive,huang2021surrogate}.
It is important to note that accurate source selection alone is insufficient to protect MSAS from negative transfer, as the identified knowledge, despite being of relatively high quality, may be unhelpful for the target task when considered from the perspective of absolute quality~\cite{xue2023solution}.
As for how to transfer, one needs to determine the method for transferring the knowledge across tasks, which can be broadly divided into three categories: non-adaptive transfer~\cite{shen2024surrogate}, mapping-based adaptation~\cite{xue2020affine}, and mixture-based adaptation~\cite{wang2024distilling}.
However, it is still unclear whether a domain adaptation method can theoretically guarantee that its adapted knowledge will always be beneficial for the target task~\cite{xue2023solution}.
As a result, reliance on transfer methods alone remains insufficient to avoid negative transfer.
Regarding when to transfer, it is crucial to determine the timing for executing knowledge transfer during the optimization process, which involves making a series of decisions about whether to transfer knowledge at each moment.
An ideal solution in this context should align the decisions of whether to transfer with the actual benefits of transferring knowledge at different moments, ensuring that nonnegative performance gain is achieved with adaptive knowledge transfer~\cite{xue2024surrogate}.
In this sense, an effective solution for determining when to transfer knowledge is crucial for developing advanced MSAS algorithms that are free from the threat of negative transfer, which will be the primary focus of this study.

In the literature, a wide variety of techniques has been developed to tackle the issue of when to transfer, which can be broadly divided into two groups: rigid timing and adaptive timing.
Prespecified transfer probability~\cite{wu2024evolution} and prespecified transfer interval~\cite{feng2018evolutionary} are two popular strategies that rigidly determine the timing for knowledge transfer.
However, pre-determining an appropriate transfer intensity that properly reflects the benefit of transferring knowledge is not trivial~\cite{xue2023solution}, particularly when the task relatedness is unknown in advance.
To alleviate this issue, several adaptive methods have been developed over the years, which can be roughly divided into two classes: a) non-empirical methods that estimate the timing for knowledge transfer based on specific a priori criteria~\cite{binh2019multi,bali2020multifactorial,cai2021evolutionary,lin2023multiobjective,zhang2021surrogate,shen2024surrogate}; and b) empirical methods that determine the timing for knowledge transfer according to its historical performance~\cite{liaw2019evolutionary,li2020multifactorial, li2023evolutionary,wu2024learning,tan2024surrogate,wang2024evolutionary}.
However, both of these two classes of methods exhibit certain limitations.
On the one hand, non-empirical methods are inherently limited due to their imprecise modeling of the benefits of transferring knowledge in complex multitasking environments.
On the other hand, empirical methods often require a sufficient number of observations to be effective, which limits their utility when empirical evidence is scarce, especially during the early stage of optimization.
Therefore, a seamless integration of these two types of methods is desirable to determine the optimal timing for knowledge transfer, enabling adaptive utilization (or suppression) of knowledge based on its helpfulness to the target task.
However, such integration has not been thoroughly explored, leaving existing MSAS algorithms vulnerable to the risk of negative transfer.
Additionally, surrogate dependency~\cite{tan2024surrogate} and incapability of many-tasking~\cite{shen2024surrogate} are two prominent issues that need to be urgently addressed in MSAS.

In light of the above, this paper proposes a novel solution to the issue of when to transfer, namely Bayesian competitive knowledge transfer (BCKT), to improve the search performance of various MSAS optimizers in a plug-and-play manner.
Firstly, transferability is defined as the improvement achieved by source knowledge\footnote{Knowledge refers to source information that can explicitly or implicitly guide the generation of candidate solutions for the target task. In this work, we consider elite solutions as transferable knowledge.} in solving the target task~\cite{xue2023solution}, compared to the promising solution found.
This quantity can be expressed through various scalarized measures, enabling estimation from a Bayesian perspective that incorporates both prior knowledge and empirical evidence.
Subsequently, by considering the promising solution on the target task and the elite solutions from the source tasks as task-solving knowledge, the estimated transferability allows them to compete, thereby enabling the identification of the winner for real evaluation.
In this way, a nonnegative performance gain for each EOP is expected to be achieved through adaptive knowledge transfer from the other EOPs acting as source tasks.
Furthermore, the proposed BCKT method seamlessly adopts a many-task formulation, allowing it to efficiently solve more than two tasks simultaneously.
Experimental studies conducted on a series of benchmark suites and a practical case study demonstrate the superiority of our BCKT method compared to the state-of-the-art.
The main contributions of this paper are as follows:

\begin{itemize}
\item The transferability is represented by a latent variable $\tau$, enabling it to be estimated from a probabilistic perspective.
Both our prior beliefs and empirical observations regarding transferability can be effectively integrated within the Bayesian inference framework, leading to more reliable estimates of transferability in complex tasks as the optimization progresses.

\item By modeling both a priori belief and posterior observation about transferability as Gaussian distributions for Bayesian estimation, we can analytically track the up-to-date distribution of transferability during the optimization process---thanks to the conjugacy of Gaussians.
Moreover, the asymptotic efficiency of such estimation theoretically guarantees that our understanding of transferability can be refined progressively as the optimization proceeds.

\item Adaptive knowledge transfer is achieved through a competition mechanism: the quality of elite solutions from source tasks on the target task is quantified based on their up-to-date transferability---so that these source solutions can compete with the most promising solution of the target task, ultimately electing the winner for real evaluation.
This enables our decisions on knowledge transfer to align with the actual benefits of those transfer, leading to nonnegative performance gain for the target optimization.

\end{itemize}

\begin{figure}[ht]
	\centering
	\includegraphics[width=3.4in]{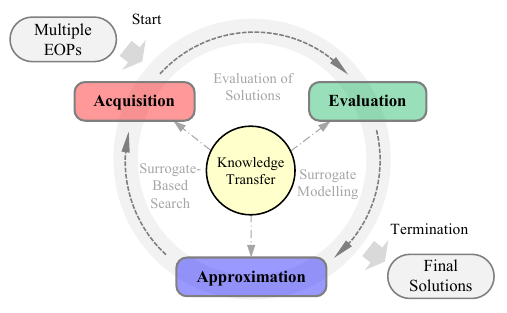}
	\caption{High-level structure of MSAS.}
	\label{fig:MSAS}
\end{figure}

The remainder of this paper is organized as follows.
Section \ref{section:pre} reviews related work on MSAS and various techniques addressing the issue of when to transfer, along with the motivations behind this study.
Section \ref{section:BCKT} presents the proposed BCKT method and the overall implementation of MSAS-BCKT.
The experimental results obtained on the benchmark suites and practical case studies are reported and analyzed in Section \ref{section:exp}.
Finally, Section \ref{section:con} concludes the paper with several prospects for future research.

%################################################################################
%######################################divider######################################
%################################################################################

\section{Related Work}
\label{section:pre}

\subsection{Multi-task Surrogate-Assisted Search (MSAS)}

Through knowledge transfer (or sharing) across multiple EOPs of interest, MSAS aims to achieve better overall optimization performance on these problems~\cite{huang2021surrogate,tan2024surrogate,shen2024surrogate}, with its high-level structure illustrated in Fig. \ref{fig:MSAS}.
Typically, MSAS involves multiple optimization processes of surrogate-assisted search, each of which is responsible for solving a particular individual EOP with three iterative steps as follows:

\begin{itemize}
\item[1)] Approximation: A joint multi-task surrogate~\cite{swersky2013multi,tan2024surrogate} or an individual surrogate model~\cite{huang2021surrogate,shen2024surrogate} is constructed (or updated) to approximate the original EOP for subsequent optimization that may require extensive evaluations.

\item[2)] Acquisition: To acquire a promising solution for the EOP of interest, an optimization algorithm is employed to optimize the surrogate model with a particular infill criterion, such as the expected improvement~\cite{jones1998efficient}.

\item[3)] Evaluation: The promising solution will be evaluated and added to the database for enriching the samples. 
\end{itemize}

In MSAS, knowledge transfer (or sharing) across the EOPs may occur in any of the three steps in Fig. \ref{fig:MSAS}, such as the kernelization of cross-task samples in multi-task surrogate modeling~\cite{tan2024surrogate}, the surrogate aggregation for solution acquisition~\cite{min2017multiproblem}, and the transfer of promising solutions for real evaluation~\cite{shen2024surrogate}, etc.
Significantly, methods of transferring promising solutions exhibit superior flexibility and scalability due to their independency of the backbone surrogate model and search mechanism adopted, making them favorably coherent with the state-of-the-art of surrogate-assisted optimization~\cite{xue2024surrogate}.

\subsection{When to Transfer}

\label{section:pre-when}

For knowledge transfer of promising solutions across EOPs, the fundamental issue of ``when to transfer'' should be adequately addressed, as improper evaluation on a target EOP with irrelevant solutions could greatly hinder its solving efficiency~\cite{wang2023recent}.
Even worse, the risk of performance degradation mentioned above will be exacerbated further by the shortage of evaluation calls, which is very common in most EOPs.
Promisingly, the above challenge can be tackled by estimating the ground-truth quality of source solutions on the target task, which is also known as \emph{transferability}~\cite{xue2023solution}.
In the literature, different scalars have been defined to represent the transferability, such as transfer probability~\cite{gupta2015multifactorial} and transfer interval~\cite{feng2018evolutionary}.
In this work, we define a latent variable $\tau$ for such representation and use it to decide whether to transfer at each particular moment, eventually leading to a comprehensive solution to ``when to transfer''.
An ideal solution that well aligns the decisions of whether to transfer with the transferability of knowledge at different moments will enable the utilization of useful knowledge with adaptive suppression of useless knowledge~\cite{xue2024surrogate}.
In this sense, the quality of a solution to ``when to transfer'' primarily comes from the accuracy of assessing the latent variable $\tau$.
A variety of techniques in this respect can be divided into three categories from an epistemological perspective: a priori methods~\cite{zhang2021surrogate,wang2021surrogate}, a posteriori approaches~\cite{li2023evolutionary,liaw2020evolution}, and hybrid ones, as reviewed separately in what follows.

\subsubsection{A Priori Methods} This class of methods assume that the transferability can be assessed without actually experiencing the process of knowledge transfer, which can be roughly divided into two categories: prior-informed prespecification and adaptive estimation based on a priori criteria.
Particularly, prespecified transfer probability~\cite{liu2018surrogate} and prespecified transfer interval~\cite{feng2018evolutionary} for triggering knowledge transfer randomly and periodically, respectively, are two most widely used prespecification strategies, whose settings heavily rely on one's prior understanding of the transferability of interest.
A prespecified small transfer interval (or large transfer probability~\cite{huang2021surrogate}) in cases of low transferability may increase the risk of negative transfer, while a small transfer probability (or large transfer interval) would diminish the potential for positive transfer across tasks with high transferability~\cite{chen2020adaptive}.
In response to this issue, a variety of adaptive methods that estimate the transferability online with certain a priori criteria have been developed, including the coefficients of mixture models~\cite{bali2020multifactorial}, similarity measurements~\cite{cai2021evolutionary,shen2024surrogate}, search state-based measures~\cite{binh2019multi,lin2023multiobjective}, etc.
However, these a priori methods are always intrinsically deficient due to their imprecise modeling of the transferability with limited understandings.
Such limitation can be largely attributed to the complex nature of the continuously evolving solutions being transferred in the multitasking environment, making it difficult to accurately assess their transferability with a priori clues only.
Besides, transfer results as empirical observations showing the transferability are always neglected by the a priori methods, making them miss out the opportunity to improve their ability of transferability assessment with the empirical observations.
In summary, a priori methods have difficulties assessing the the transferability accurately due to the intrinsically complex nature of the continuously evolving solutions shared across different tasks.

\subsubsection{A Posteriori Approaches} This category of methods are grounded in observations of the transferability reflected by transfer results~\cite{xu2021evolutionary,wu2024multitasking}, which can be divided into two classes: feedback-based metrics and learning-based predictors.
In essence, all the feedback-based metrics are grounded in a common principle: the unknown transferability can be gradually ascertained as we gather more observations~\cite{liaw2020evolution,lin2024multiobjective}, despite minor variations in their measurement formulations.
Several representative examples in this respect can be found in~\cite{liaw2019evolutionary,li2020multifactorial}.
By contrast, learning-based predictors typically assume that the transferability is learnable with a finite number of observations and the learned model can be used to predict future decisions, such as the supervised classification of positive and negative transfers in~\cite{lin2019multiobjective}, multi-task or transfer Gaussian process models\footnote{Please note that while these transfer GPs are considered a posteriori in approximating optimization functions, the evidence they use does not directly address the issue of ``when to transfer.''}~\cite{tan2024surrogate,wang2024evolutionary}, as well as the reinforcement learning for transfer intensity adjustment in~\cite{li2023evolutionary,wu2024learning}.
Although these a posteriori methods are empirically faithful to the transfer results, they always require an adequate number of observations to obtain a satisfactory estimate of the transferability.
However, transfer observations at the early stage of optimization are very limited, which would weaken the a posteriori methods' effectiveness greatly.
In a word, a posteriori methods could be inefficient in assessing the transferability due to their passive perception of the transferability that solely relies on the empirical evidence.

\subsubsection{Hybrid Methods} 
\begin{figure}[ht]
	\centering
	\includegraphics[width=3.5in]{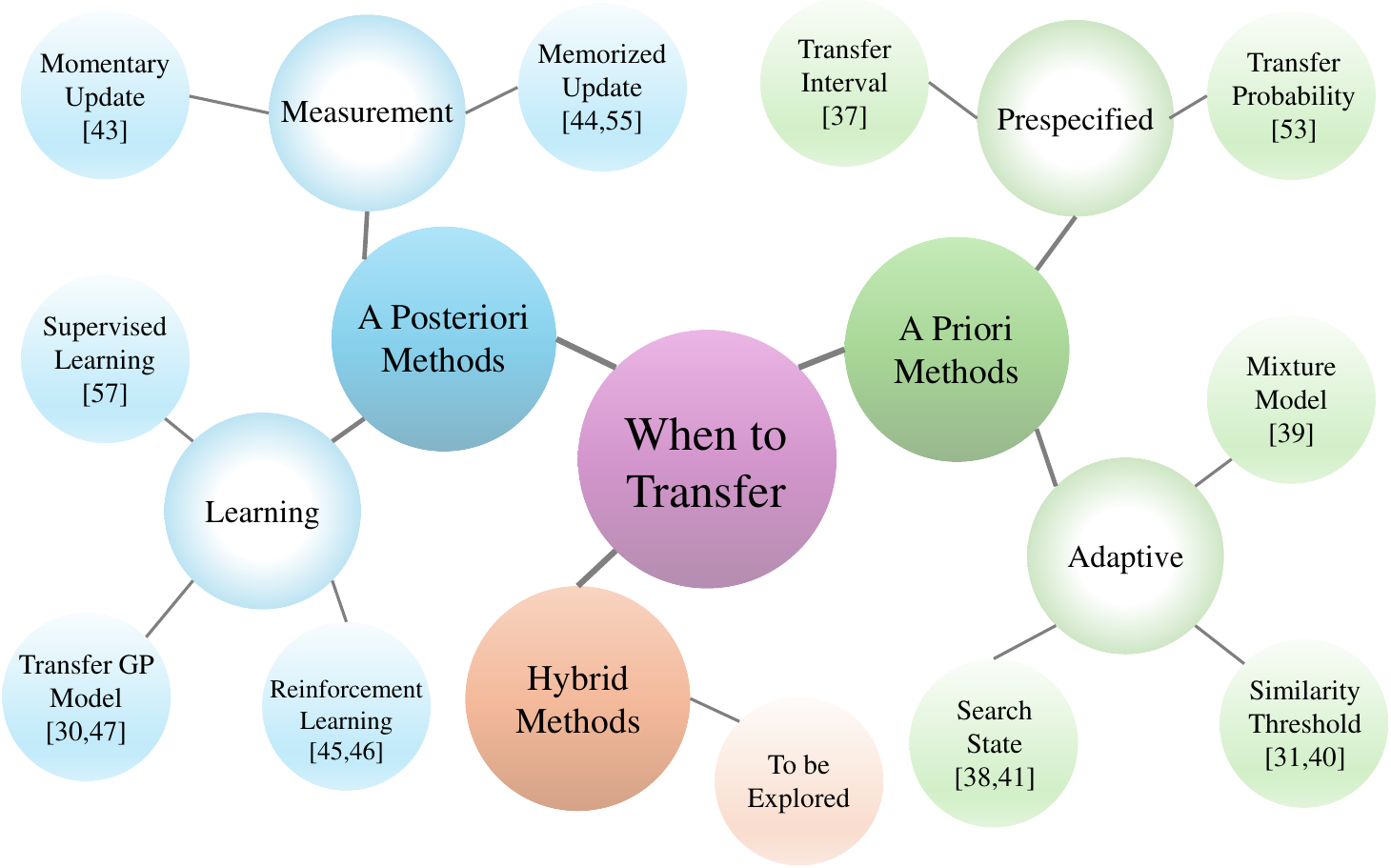}
	\caption{Taxonomy of the existing techniques for when to transfer: an epistemological perspective.}
	\label{fig:works}
\end{figure}

According to the analyses above, one can find that a priori and a posteriori methods exhibit different advantages.
On the one hand, a priori methods can offer certainty and foundational insights if the prior understandings involved can correctly lead to the manifestation of the transferability, enabling their outstanding efficiency of transferability assessment.
On the other hand, a posteriori methods can guarantee their empirical relevance to the transferability and improve with new observations over time, making them more applicable to complex situations in which important nuances might be missed by a priori reasoning.
Therefore, it is natural to integrate these two types of methods to achieve a more accurate assessment of the transferability.
However, existing hybrid methods primarily focus on source selection without adequately addressing the issue of when to transfer, as done in~\cite{chen2020adaptive,huang2021surrogate}, which combine the Kullback–Leibler divergence with a feedback-based reward metric.
Consequently, the prespecified knowledge transfer in ~\cite{chen2020adaptive,huang2021surrogate} may cause negative transfer in cases of entirely low transferability (i.e., the knowledge from all the source tasks is unhelpful for the target task), even if the source selection is very accurate.
This situation would deteriorate further when solving EOPs with limited evaluation calls available, as correcting misspecified timing of knowledge transfer with an adequate number of observations is computationally unaffordable.
To summarize, a holistic taxonomy of the techniques addressing the issue of when to transfer is presented in Fig. \ref{fig:works}.
In conclusion, there is an urgent need for developing a hybrid method that can assess the transferability comprehensively with both a priori and a posteriori clues to decide when to transfer across EOPs---so that adaptive knowledge transfer can be favorably achieved.

% However, the combination in~\cite{chen2020adaptive} is made in an ad hoc fashion without rigorous analyses on how it can capture the unknown transferability.

To sum up, the existing MSAS methods suffer from three inadequacies when solving EOPs, as itemized in what follows.

\begin{enumerate}

    \item \emph{Model dependency}: The MSAS method in~\cite{tan2024surrogate} is undetachably bonded to the associated Bayesian surrogate, making it inevitably inherit the model's downsides~\cite{xue2024surrogate}. Besides, the model dependency of~\cite{tan2024surrogate} makes its  transfer method disconnected from general SAS algorithms.
    \item \emph{Poor capability of many-tasking}: The MSAS methods in~\cite{shen2024surrogate,tan2024surrogate} have no many-tasking instantiations, making them incapable of handling more than two optimization tasks simultaneously.  
    \item \emph{High risk of negative transfer}: The MSAS algorithm in~\cite{huang2021surrogate} transfers knowledge all the time and the method in~\cite{shen2024surrogate} employs a similarity threshold to decide whether to transfer, both of which ignore transfer results reflecting the true transferability, making them troubled by the risk of negative transfer.
\end{enumerate}

\subsection{Motivations}

In light of the above, we aim to develop a knowledge transfer method to improve MSAS, which is supposed to exhibit the following three favorable merits:
\begin{enumerate}
    \item \emph{Portability}: The knowledge transfer method is independent of surrogate models, enabling it to be coherent with the state-of-the-art in SAS~\cite{xue2024surrogate}.
    \item \emph{Scalability}: It should be scalable in terms of the number of EOPs to be solved concurrently, allowing for efficient many-task optimization.
    \item \emph{Adaptability}: It should be able to pursue positive transfer while avoiding the risk of negative transfer aptly. Such adaptive knowledge transfer holds much significance for solving EOPs with very limited evaluation calls.
\end{enumerate}

Moreover, we are interested in the capability of our method to estimate knowledge transferability, which will be firmly supported by its asymptotic unbiasedness.

%################################################################################
%######################################divider######################################
%################################################################################

\section{Proposed Methods}

\label{section:BCKT}

\begin{figure}[ht]
	\centering
	\includegraphics[width=3.5in]{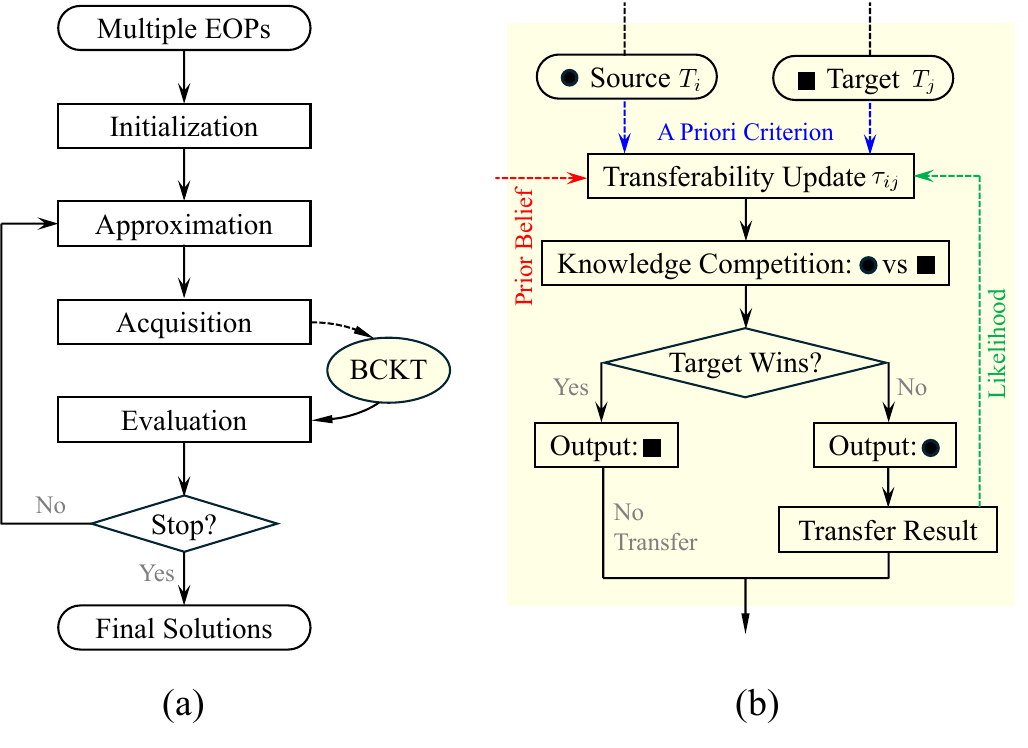}
	\caption{Flowcharts of MSAS-BCKT and BCKT: (a) the overall structure of MSAS-BCKT; (b) the workflow of BCKT.}
	\label{fig:framework}
\end{figure}

Fig. \ref{fig:framework} demonstrates the overall structure of MSAS-BCKT and the detailed workflow of BCKT.
From Fig. \ref{fig:framework}(a), we can see that MSAS-BCKT differs from the basic MSAS due to the incorporation of BCKT, which enables adaptive knowledge transfer during the evaluation phase.
Specifically, the proposed BCKT in this work is extended from the concept of competitive knowledge transfer (CKT) in~\cite{xue2024surrogate}.
However, unlike CKT, which relies solely on a priori similarity to evaluate the transferability through inductive analogy, BCKT approaches this assessment from a Bayesian perspective.
As shown in Fig. \ref{fig:framework}(b), the transferability in BCKT is dynamically updated based on both a priori and a posteriori information throughout the optimization process.
The former may consist of specific prior beliefs or criteria, while the latter is derived from actual transfer results that reflect the likelihood of transferability.
Subsequently, the updated transferability is used to estimate the quality of source elite solutions on the target task.
This enables the identification of the winner between the source elite solutions and the target candidate solution for real evaluation, thereby facilitating adaptive knowledge transfer for enhanced optimization performance.
In what follows, the formal definition of transferability in the context of CKT is presented first.
Afterward, the core of BCKT---the Bayesian inference of transferability---is introduced in detail.
Finally, the implementation of BCKT within MSAS (i.e., MSAS-BCKT) is presented, accompanied by a thorough analysis of its theoretical efficacy and computational complexity.

\subsection{Transferability in CKT}

In CKT, two central concepts---\emph{internal improvement} and \emph{external improvement}---are defined to represent the quality of source and target knowledge in relation to the target problem-solving~\cite{xue2024surrogate}.
When transferring source knowledge in the form of promising solutions, these two improvements can be explicitly defined as follows:

\begin{figure}[ht]
	\centering
	\includegraphics[width=3.5in]{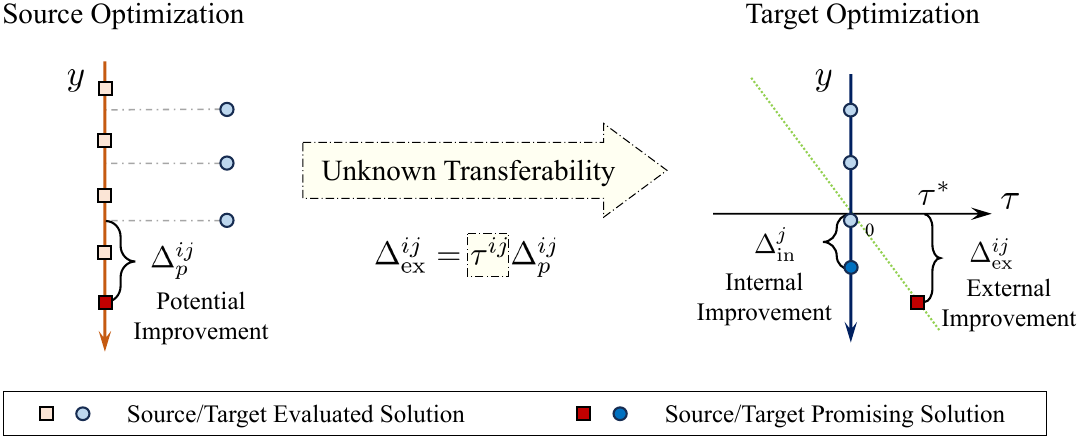}
	\caption{Illustration of the internal and external improvements.}
	\label{fig:transferability}
\end{figure}

\begin{itemize}
\item{Internal Improvement}: From the perspective of a single target task, it refers to the estimated improvement of a promising solution obtained through SAS, based on a specific infill criterion.
\item{External Improvement}: It represents the estimated improvement in target optimization achieved by transferring a promising solution\footnote{Without loss of generality, we consider the best solution found for transfer. Alternatively, one may define their own ``promising'' solutions based on sophisticated infill criteria.} from a source task.
\end{itemize}

In a multi-task environment, the internal improvement of the $j$-th task, serving as the target, can be estimated as follows:
\begin{equation}
\Delta^{j}_{\mathrm{in}} = \min_{\boldsymbol{x} \in X^{j}} \mathcal{I}(\boldsymbol{x}) - \mathcal{I}(\boldsymbol{x}^{j}_p),
\label{eq:imp_in}
\end{equation}where $\Delta_{\mathrm{in}}^j$ denotes the internal improvement of the $j$-th task, $\mathcal{I}\left(\cdot\right)$ represents the infill criterion (i.e., acquisition function) to be minimized on the target surrogate, $X^j$ represents the evaluated solutions stored in the $j$-th task's database $\mathcal{D}^j$, $\mathcal{D}^j=\lbrack X^j,\boldsymbol{y}^j\rbrack$ contains both evaluated solutions $X^j$ and their real objective values $\boldsymbol{y}^j$, $\boldsymbol{x}_p^j$ denotes the most promising solution obtained by SAS for the $j$-th task, $n$ is the number of tasks.

To estimate the external improvement, the original CKT method relies exclusively on analogical reasoning, using rank correlation to assess source-target similarity~\cite{xue2024surrogate}.
The greater the similarity between source and target tasks, the more confident we are in transferring promising solutions to enhance search efficiency~\cite{xue2025theoretical}.
However, a priori analogy struggles to effectively capture the transferability of source solutions in a multi-task environment due to their inherent complexity and dynamic nature.
This challenge motivates us to integrate transferability directly into the external improvement, enabling it to account for both a priori insights and posterior evidence.
Considering the $i$-th task as the source and the $j$-th task as the target, the external improvement is estimated as follows:
\begin{equation}
\begin{split}
\Delta^{ij}_{\mathrm{ex}}&=\tau^{ij}\Delta_p^{ij}\\
&=\tau^{ij}\left[ \min_{\boldsymbol{x} \in X^j}\hat{f}_i(\boldsymbol{x}) - \min{(\boldsymbol{y}^i)} \right]\frac{\max(\boldsymbol{y}^j)}{\max(\boldsymbol{y}^i)},
\end{split}
\label{eq:imp_ex_our}
\end{equation}
where $\Delta^{ij}_{\mathrm{ex}}$ represents the external improvement achieved by the best-ever-found solution of the $i$-th task for the $j$-th task, $\tau^{ij}\in\mathbb{R}$ denotes a latent variable that reflects the transferability of the best-ever-found solution from the $i$-th task to the $j$-th task, $\Delta_p^{ij}$ represents the improvement of the source best solution relative to all the target solutions, evaluated from the perspective of the source task, $\hat{f}_i(\cdot)$ denotes the source surrogate model.

To elaborate, we compare the internal and external improvements in Fig. \ref{fig:transferability}.
First, the improvement of the source promising solution relative to all the target solutions is estimated using the source surrogate model.
This improvement is then utilized to calculate the external improvement, factoring in the source-target transferability as described in Eq. \eqref{eq:imp_ex_our}.
Taking $\tau$ as the x-axis (as shown on the right side of Fig. \ref{fig:transferability}), the green dotted line, which represents the calculation of $\Delta_{\mathrm{ex}}$, has a slope of $\Delta_p$.
Finally, with the estimated external improvement, adaptive knowledge transfer can be achieved as follows:
\begin{equation}
\boldsymbol{x}^j_{\mathrm{eva}} = 
\begin{cases} 
    \boldsymbol{x}^j_p, & \text{if}\,\Delta_{\mathrm{in}}^{j} \ge \Delta_{\mathrm{ex}}^{ij}\,\left(\mathrm{no\,\,transfer}\right),\\
    \boldsymbol{x}^i_{\mathrm{best}}, & \text{if}\,\Delta_{\mathrm{in}}^{j} < \Delta_{\mathrm{ex}}^{ij}\,\left(\mathrm{knowledge\,\,transfer}\right),
\end{cases}
\label{eq:ckt}
\end{equation}
where $\boldsymbol{x}^j_{\mathrm{eva}}$ denotes the solution to be evaluated by the real function of the $j$-th task, $\boldsymbol{x}^i_{\mathrm{best}}$ represents the best-ever-found solution of the $i$-th task.
It is evident that determining the unknown transferability $\tau^{ij}$ in $\Delta_{\mathrm{ex}}^{ij}$ is essential for adaptive knowledge transfer in Eq. \eqref{eq:ckt}.

\subsection{Bayesian Inference of Transferability}

Recognizing the difficulty in accurately determining the transferability of continuously evolving elite solutions between two tasks based solely on a priori clues\footnote{In oracle-based optimization, the ground-truth quality of any solution can only be determined through real evaluation. Furthermore, a single observation is insufficient to ascertain the transferability of rich knowledge between tasks.}, we propose treating the transferability representation $\tau$ in Eq. \eqref{eq:imp_ex_our} as an deterministic yet unknown latent variable.
Its true value $\tau_*$ is supposed to be gradually approximated as the optimization unfolds.
Specifically, we utilize Bayesian inference to estimate $\tau$ due to its flexibility in integrating both prior and posterior information, as given by
\begin{equation}
p\left(\tau\mid {T}\right) = \frac{p\left({T}\mid\tau\right) p\left(\tau\right)}{p\left({T}\right)},
\label{eq:Bayesian}
\end{equation}
where $p\left(\tau\right)$ denotes the prior distribution of $\tau$, ${T}$ represents an observation of $\tau$, $p\left({T}\mid\tau\right)$ represents the sampling distribution of ${T}$ conditional on $\tau$, which is also known as likelihood, $p\left({T}\right)$ denotes the marginal distribution of $T$ marginalized over $\tau$, $p\left(\tau\mid {T}\right)$ denotes the posterior distribution of $\tau$ after taking into account both the prior and likelihood information.

The prior distribution $p(\tau)$ in Eq. \eqref{eq:Bayesian} can be interpreted as a pooled collection of prior beliefs~\cite{johnson2010methods}, capable of incorporating two types of information: data-free prior belief $\textcolor{red}{p(\tau)}$ and a priori data-based transferability quantification criteria $\textcolor{blue}{p^c(\tau)}$ (e.g., population-based similarity), as illustrated by the red and blue arrows in Fig. \ref{fig:framework}(b), respectively.
Prior to knowledge transfer, prior belief $\textcolor{red}{p(\tau)}$ may originate from white-box insights into the transferability, while they can be designated as the previous posterior obtained from the preceding transfer.
In this study, we do not delve into domain-specific prior beliefs, but in practice, such beliefs should be utilized if they provide relevant information about $\tau_*$.
For both the prior and likelihood terms, we propose modeling them as Gaussians.
This ensures that the posterior distribution, $p\left(\tau\mid{T}\right)$, remains Gaussian due to conjugacy properties.
As a result, the inference maintains analytical tractability and avoids significant computational overhead.
Upon the occurrence of the $k$-th knowledge transfer from the $i$-th task to the $j$-th task, the distribution of $\tau^{ij}$ can be updated as follows:
\begin{equation}
p^{({k})}(\tau^{ij}) \propto \textcolor{green}{p(T^{ij}_{k}\mid\tau^{ij})}\textcolor{blue}{p^{c}(\tau^{ij})}\textcolor{red}{p^{({k}-1)}(\tau^{ij})},
\label{eq:latest_tau}
\end{equation}
where $p^{({k})}(\tau^{ij})$ is the posterior distribution of $\tau^{ij}$ following ${k}$ observations of knowledge transfer.
For clarity, we use colors to highlight the three terms in Eq. \eqref{eq:latest_tau}, aligning them with the three sources of transferability information illustrated in Fig. \ref{fig:framework}(b), each of which is detailed below.

\subsubsection{Prior Belief}
In Bayesian inference, the prior belief serves as the starting point for updating our understanding when new observations become available.
In the absence of such priors when $k=1$, it is recommended to use a non-informative constant value, i.e., $\textcolor{red}{p^{(0)}(\tau^{ij})}=1$.
For $k\ge2$, $\textcolor{red}{p^{(k-1)}(\tau^{ij})}$ can be designated as the latest posteriori.

\subsubsection{A Priori Criterion}
Without loss of generality, we employ the Spearman's rank correlation coefficient\footnote{Any other a priori criteria can be leveraged in the same way with proper uncertainty quantification.} as the prior assessment of $\tau^{ij}$~\cite{xue2024surrogate}, as given by
\begin{equation}
R^{ij}_{k}=\frac{\mathrm{cov}(\mathcal{R}\lbrack\hat{f}_i(\boldsymbol{x})\mid_{\boldsymbol{x} \in X^j}\rbrack,\mathcal{R}\lbrack\boldsymbol{y}^j\rbrack)}{\mathrm{std}(\mathcal{R}\lbrack\hat{f}_i(\boldsymbol{x})\mid_{\boldsymbol{x} \in X^j}\rbrack)\mathrm{std}\left(\mathcal{R}\lbrack\boldsymbol{y}^j\rbrack\right)},
\label{eq:srrc}
\end{equation}
where $R^{ij}_{k}$ denotes the rank correlation between the $i$-th task as the source and the $j$-th task as the target, $\mathcal{R}\lbrack \boldsymbol{a}\rbrack$ is the rank vector of $\boldsymbol{a}$.
Then, we can interpret this calculated correlation coefficient as a prior assessment of $\tau^{ij}$, there leading us to

\begin{equation}
\textcolor{blue}{p^{c}(\tau^{ij})}=\mathcal{N}(R^{ij}_{k},\sigma^2_{R_{k}}),
\label{eq:prior_Gaussian}
\end{equation}
where $\mathcal{N}(\mu,\sigma^2)$ represents the univariate Gaussian with mean $\mu$ and variance $\sigma^2$, $\sigma^2_{R_{k}}$ denotes the uncertainty of the rank correlation in capturing $\tau^{ij}_*$ during the $k$-th knowledge transfer.
\subsubsection{Likelihood}
According to Eq. \eqref{eq:imp_ex_our}, the observation $T^{ij}_{k}$ in the likelihood term can be calculated as follows:
\begin{equation}
T^{ij}_{k}=\frac{\Delta_\mathrm{ex}^{ij}}{\Delta_p^{ij}}=\frac{\min{(\boldsymbol{y}^j)}-f_j(\boldsymbol{x}_{\mathrm{best}}^i)}{\Delta_p^{ij}},
\label{eq:obsevaton_single}
\end{equation}
where $f_j(\cdot)$ represents the objective function of the $j$-th task.
Then, we have the likelihood as follows:

\begin{equation}
\textcolor{green}{p(T^{ij}_{k}\mid\tau^{ij})}=\mathcal{N}(T^{ij}_{k},\sigma^2_{T_{k}}),
\label{eq:likelihood_Gaussian}
\end{equation}
where $\sigma^2_{T_{k}}$ denotes the uncertainty of $T^{ij}_{k}$ in reflecting $\tau^{ij}_*$ during the $k$-th knowledge transfer.

With the conjugacy of Gaussians, the posterior distribution in Eq. \eqref{eq:latest_tau} can be analytically obtained as follows:
\begin{equation}
\begin{split}
p^{({k})}(\tau^{ij})&\propto\prod_{1\le l\le k}\mathcal{N}(T^{ij}_l,\sigma^2_{T_l})\cdot\mathcal{N}(R^{ij}_l,\sigma^2_{R_l})\\
&=\mathcal{N}(\hat{\tau}_{k}^{ij},\sigma_{k}^2),
\end{split}
\label{eq:latest_tau_explicit}
\end{equation}
where
\begin{equation}
\hat{\tau}_{k}^{ij}=\omega\bigg\lbrack\sum_{1\le l\le k}\frac{T^{ij}_l}{\sigma^2_{T_l}}\bigg\rbrack\sigma^2_{T}+\left(1-\omega\right)\bigg\lbrack\sum_{1\le l\le k}\frac{R^{ij}_l}{\sigma^2_{R_l}}\bigg\rbrack\sigma^2_{R},
\label{eq:tau_mu}
\end{equation}
and
\begin{equation}
\sigma^2_{k}=\frac{\sigma^2_{T}\sigma^2_{R}}{\sigma^2_{T}+\sigma^2_{R}},
\label{eq:tau_sigma}
\end{equation}
where $\sigma^2_T=\big\lbrack\sum_{1\le l\le k}\sigma^{-2}_{T_l}\big\rbrack^{-1}$, $\sigma^2_R=\big\lbrack\sum_{1\le l\le k}\sigma^{-2}_{R_l}\big\rbrack^{-1}$, and $\omega=\sigma^2_R/(\sigma^2_T+\sigma^2_R)$.
Regarding the variances, $\sigma^{2}_{R_l}$ is set to a constant value $\sigma^{2}_{I}$ due to the rank correlation's inherent impreciseness in capturing $\tau^{ij}_*$, while $\sigma^{2}_{T_l}$ is defined as a decreasing value $\sigma^2_Ie^{-l}$, based on an assumption that the transferability is better reflected as the optimization progresses. %where $t_l$ denotes the time at which the $l$-th knowledge transfer occurs, as represented by the number of evaluated solutions $|\mathcal{D}^j|$
Then, we can rewrite Eq. \eqref{eq:tau_mu} and Eq. \eqref{eq:tau_sigma} as follows:
\begin{equation}
\hat{\tau}_{k}^{ij}=\omega\sum_{1\le l\le k} z_l T^{ij}_l+(1-\omega)\frac{1}{k}\sum_{1\le l\le k} R^{ij}_l,
\label{eq:tau_mu2}
\end{equation}
and
\begin{equation}
\sigma^2_{k}=\frac{\sigma^2_I}{{k}+\varepsilon},
\label{eq:tau_sigma2}
\end{equation}
where $\omega=\varepsilon/({k}+\varepsilon)$, $z_l=e^l/\varepsilon$, $\varepsilon=\sum_{1\le l\le k} e^l$.

After $k$ observations of knowledge transfer from the $i$-th task to the $j$-th task, posterior sampling over $p^{({k})}(\tau^{ij})$ can be performed to estimate $\tau^{ij}$, facilitating the calculation of the external improvement in Eq. \eqref{eq:imp_ex_our}, as formally given by
\begin{equation}
\tilde{\tau}^{ij}\sim\mathcal{N}(\hat{\tau}_{k}^{ij},\sigma_{k}^2),
\label{eq:sampling}
\end{equation}
where $\tilde{\tau}^{ij}$ denotes a posteriori sample that reflects our latest understanding of $\tau^{ij}$.
More generally, we can apply Eq. \eqref{eq:sampling} to all the task pairs in a multi-task EOP, ultimately leading to a matrix of inter-task transferabilities:
\begin{equation}
\ \tilde{\boldsymbol{\tau}} = 
\begin{bmatrix}
\circ & \tilde{\tau}^{12} & \cdots & \tilde{\tau}^{1n} \\
\tilde{\tau}^{21} & \circ & \cdots & \tilde{\tau}^{2n} \\
\vdots & \vdots & \ddots & \vdots \\
\tilde{\tau}^{n1} & \tilde{\tau}^{n2} & \cdots & \circ
\end{bmatrix},
\label{eq:tau_matrix}
\end{equation}
where $\tilde{\boldsymbol{\tau}}$ represents the transferability matrix, $\circ$ denotes a null placeholder.
It is important to note that $\tilde{\boldsymbol{\tau}}$ is typically asymmetric, meaning that $\tilde{\tau}^{ij}$ and $\tilde{\tau}^{ji}$ cannot be used interchangeably.
In many cases, $T_a$ benefits from $T_b$, while $T_b$ may not help $T_a$, as will be empirically demonstrated in our experiments later.

\subsection{Bayesian CKT for MSAS}

Using the transferability matrix from Eq. \eqref{eq:tau_matrix} to calculate the external improvement in Eq. \eqref{eq:imp_ex_our}, along with the internal improvement in Eq. \eqref{eq:imp_in}, we can define a unified improvement matrix as follows:

\begin{algorithm}
	\SetKwInOut{Input}{Input}
	\SetKwInOut{Output}{Output}
        \SetNoFillComment
	\caption{MSAS-BCKT}
         \label{alg:le}
	\begin{small}
		\Input{$\lbrace f_1,...,f_n\rbrace$---$n$ EOPs to be solved concurrently; $\mathrm{FE}_{\mathrm{max}}$---maximum number of real evaluations.}
		\Output{$\boldsymbol{x}^j_{\mathrm{best}}\mid_{1\le j\le n}$---best solutions.}
        \begin{tikzpicture}[remember picture,overlay]
    \draw [fill=myboxcolor, ultra thin] ($(pic cs:a) + (-1.5,0.29)$) rectangle ($(pic cs:b)+(7.3,0.34)$);
	\end{tikzpicture}
$\mathcal{D}^j\mid_{1\le j\le n}\leftarrow$ Initialize with sampling-based real evaluation;\\
$\mathrm{FE}_{\mathrm{used}}\leftarrow \sum_{j=1}^n\lvert\mathcal{D}^j\rvert$;\\
\While{$\mathrm{FE}_{\mathrm{used}}< \mathrm{FE}_{\mathrm{max}}$}
{
		\For{$j=1\to n$}
		{
                \tcc{Internal Improvement}
			$\lbrack \boldsymbol{x}^j_p, \Delta^j_{\mathrm{in}}\rbrack \leftarrow \mathtt{SAS}(\mathcal{D}^j)$;\\
		}
		\For{$j=1\to n$}
		{
			\tikzmark{a}\uIf{$\mathtt{BCKT}$ is allowed, i.e., $\mathtt{mod}(\lvert\mathcal{D}^j\rvert,\delta)=0$}
			{
                    \tcc{External Improvements}
                    \For{$i=1\to n\land i\ne j$}
                    {
                        $\tilde{\tau}^{ij}\leftarrow\mathcal{N}(\hat{\tau}_k^{ij},\sigma_k^2)$;\\
                        $\Delta^{ij}_{\mathrm{ex}}\leftarrow$ Eq. \eqref{eq:imp_ex_our} with $\tau^{ij}=\tilde{\tau}^{ij}$;\\
                    }
				\tcc{Knowledge Competition}
                    $\boldsymbol{x}^j_{\mathrm{eva}}\leftarrow\boldsymbol{x}^j_{\mathrm{win}}$ in Eq. \eqref{eq:bckt};\\
				\tcc{Transferability Update}
                    \If{$\boldsymbol{x}^j_{\mathrm{eva}}\ne\boldsymbol{x}^j_p$}
						{
							$p(\tau^{ij})\leftarrow$ Eq. \eqref{eq:latest_tau};\\
						}
			}\tikzmark{b}
			\Else
			{
				$\boldsymbol{x}^j_{\mathrm{eva}}\leftarrow\boldsymbol{x}^j_p$;\\
			}
            $\mathcal{D}^j\leftarrow\mathcal{D}^j\cup\lbrack\boldsymbol{x}^j_{\mathrm{eva}},f_j(\boldsymbol{x}^j_{\mathrm{eva}})\rbrack$;\\
            $\mathrm{FEs}_{\mathrm{used}}=\mathrm{FEs}_{\mathrm{used}}+1$;\\
		}
}
\textbf{end while}

\textbf{return} $\boldsymbol{x}^j_{\mathrm{best}}\in\mathcal{D}^j\mid_{1\le j\le n}$;
	\end{small}
\end{algorithm}

\begin{equation}
\ \boldsymbol{\Delta} = 
\begin{bmatrix}
\Delta^{11} & \Delta^{12} & \cdots & \Delta^{1n} \\
\Delta^{21} & \Delta^{22} & \cdots & \Delta^{2n} \\
\vdots & \vdots & \ddots & \vdots \\
\Delta^{n1} & \Delta^{n2} & \cdots & \Delta^{nn}
\end{bmatrix},
\label{eq:imptotal}
\end{equation}
where $\Delta^{jj}\mid_{1\le j\le n}$ denotes the internal improvement of the $j$-th task, $\Delta^{ij}\mid_{\forall i\ne j}$ represents the estimated external improvement provided by the best-ever-found solution from the $i$-th task (the source) for the $j$-th task acting as the target.
Subsequently, the promising solution obtained by SAS for a specific task can compete with the best-ever-found solutions from the remaining $n-1$ tasks to select the winning solution for real evaluation.
It is evident that knowledge transfer occurs only when the promising solution is surpassed by any of the best-ever-found solutions.
This process is referred to as Bayesian CKT (BCKT), as the external improvement is estimated from a Bayesian perspective, which can be formally formulated by
\begin{equation}
\boldsymbol{x}^j_{\mathrm{win}}=\lbrace\boldsymbol{x}^{i^*\rightsquigarrow j}\mid i^*=\underset{1\le i\le n}{\mathrm{arg\,max\,\,}}\Delta^{ij}\rbrace,\,\,\,1\le j\le n,
\label{eq:bckt}
\end{equation}
where $\boldsymbol{x}^j_{\mathrm{win}}$ denotes the winning solution to be evaluated on the $j$-th task, $\boldsymbol{x}^{i^*\rightsquigarrow j}$ represents the promising solution $\boldsymbol{x}_p^j$ when $i^*=j$ and denotes the best-ever-found solution of the $i$-th task when $i^*\ne j$.
Once knowledge transfer from the $i$-th task to the $j$-th task is permitted, the posterior distribution of $\tau^{ij}$ will be updated accordingly, as specified in Eq. \eqref{eq:latest_tau}.

The pseudo code of MSAS-BCKT is presented in Algorithm 1, with the BCKT component highlighted in yellow to differentiate MSAS-BCKT from traditional SAS.
Similar to the implementation of CKT in~\cite{xue2024surrogate}, a predefined interval $\delta$ in terms of real evaluation is used to control the occurrence of BCKT in line 7, effectively reducing frequent calculations for contiguous time slots with insignificant changes.
When the BCKT process is applied to the $j$-th task as the target, the external improvements from the remaining $n-1$ tasks, serving as the sources, are first estimated using Eq. \eqref{eq:imp_ex_our} based on the latest transferability.
Subsequently, with the estimated improvements, the target promising solution competes with the best-ever-found solutions from the sources to select a winning solution for evaluation on the $j$-th task, as shown in line 11.
Lastly, the posterior distribution of $\tau^{ij}$ will be updated based on the occurrence of knowledge transfer from the $i$-th task.
Concerning the initial distribution $p^{(0)}(\tau^{ij})$ in line 9, we can configure it based on some prior beliefs that can reflect the transferability before optimization, which is particularly common in white-box scenarios where valuable insights are available.
In cases of black-box optimization, the distribution can be initialized using an a priori data-driven transferability criterion, such as the surrogate-based correlation in Eq. \eqref{eq:srrc}.
Notably, the plug-and-play nature of BCKT enables seamless integration with any SAS solver, ensuring compatibility with future advancements in SAS.

\subsection{Theoretical Analyses of BCKT}

In this part, we begin by validating the theoretical effectiveness of BCKT through an examination of the asymptotic unbiasedness and efficiency of transferability estimation.
Then, the computational complexity of BCKT is analyzed in detail.

\subsubsection{Asymptotic Unbiasedness and Efficiency} The posterior mean in Eq. \eqref{eq:tau_mu2} is asymptotically unbiased under the squared error risk~\cite{lehmann2006theory}:
\begin{equation}
\sqrt{k}\left(\hat{\tau}^{ij}_k-\tau^{ij}_*\right)\to\mathcal{N}(0,I^{-1}_*),
\label{eq:aue}
\end{equation}
where $I_*$ denotes the Fisher information of $\tau^{ij}_*$.
The unbiasedness in Eq. \eqref{eq:aue} can also be roughly obtained by examining $\hat{\tau}^{ij}_k$ when $k\to\infty$, which is given by
\begin{equation}
\lim_{k\to\infty}\hat{\tau}^{ij}_k=\lim_{\omega\to1}\hat{\tau}^{ij}_k=\sum_{1\le l\le k} z_l T^{ij}_l.
\label{eq:tau_inf}
\end{equation}
This simply indicates that the impact of imprecise transferability modeling, based on any a priori criterion (specifically $R^{ij}$ in our case), will be gradually washed out as we accumulate more real observations of $\tau^{ij}_*$.
Additionally, we are interested in the efficiency of BCKT in approaching $\tau^{ij}_*$---how quickly can BCKT determine the true transferability such that its estimated external improvements are accurate?
The answer to this question is the Fisher information $I_*$, which quantifies the amount of information that $R^{ij}$ and $T^{ij}$ carry about $\tau^{ij}_*$, as can be explicitly given by
\begin{equation}
I_*=\textcolor{green}{I_T}+\textcolor{blue}{I_R}=\sum_{1\le l\le k}\textcolor{green}{\sigma_{T_l}^{-2}}+\textcolor{blue}{\sigma_{R_l}^{-2}}=\frac{k+\varepsilon}{\sigma_I^2}.
\label{eq:fisher}
\end{equation}
It simply shows that a larger Fisher information with improved asymptotic efficiency can arise from either an increase in the number of observations (i.e., a larger $k$) or a reduction in the uncertainties of $T$ and $R$.
This makes intuitive sense---A low uncertainty of $R$ suggests its high reliability in transferability estimation, while a low uncertainty of $T$ indicates that $\tau^{ij}_*$ can be accurately estimated with fewer real observations, both of which contribute to enhanced asymptotic efficiency.
More generally, the Fisher information can account for data-free prior beliefs $B$ as follows:
\begin{equation}
I_*=\textcolor{green}{I_T}+\textcolor{blue}{I_R}+\textcolor{red}{I_B}.
\label{eq:fisher_general}
\end{equation}
This result is intuitively graspable---the asymptotic efficiency can be effectively enhanced by integrating diverse data sources that carry varying amounts of information about $\tau^{ij}_*$.
From a high-level perspective, this study on the hybridization of a priori and a posteriori methods is fundamentally motivated by the implications of Eq. \eqref{eq:fisher_general}.

\subsubsection{Computational Complexity}

\begin{table}[ht]
	\caption{Wins, ties and losses of MSAS-BCKT against MSAS with the 6 backbone optimizers on the 9 MTOPs.}
	\centering
	\footnotesize
	\heavyrulewidth=0.12em
	\lightrulewidth=0.1em
	\cmidrulewidth=0.1em
	\setlength\tabcolsep{3pt}
	\begin{tabular}{lcccc}
		\toprule
Optimizer&HS Problems&MS Problems&LS Problems&Summary\\
		\midrule
		BO-LCB~\cite{shahriari2016taking}&2/4/0&2/4/0&0/6/0&4/14/0\\
		\midrule
		TLRBF~\cite{li2022three}&3/3/0&2/4/0&0/6/0&5/13/0\\
		\midrule
		GL-SADE~\cite{wang2023surrogate}&2/4/0&2/4/0&1/5/0&5/13/0\\
		\midrule
		DDEA-MESS~\cite{yu2022data}&2/4/0&2/4/0&0/5/1&4/13/1\\
		\midrule
		LSADE~\cite{jakub2023combining}&3/3/0&3/3/0&0/6/0&6/12/0\\
		\midrule
		AutoSAEA~\cite{xie2024surrogate}&2/4/0&3/3/0&0/6/0&5/13/0\\
		\midrule
		w/t/l &14/22/0&14/22/0&1/34/1&29/78/1\\
		\bottomrule
	\end{tabular}
	\label{tab:multi-table}
\end{table}

As shown in Algorithm 1, the computational complexity of BCKT primarily arises from the calculation of the external improvements in line 10, while the remaining calculations are computationally efficient due to their analytical forms.
Specifically, the overall complexity of calculating the external improvements for $n$ EOPs in each BCKT process is $O\left(n(n-1)\cdot\mathrm{FEs}_{\mathrm{used}}\cdot c_\mathrm{p}\right)$, where $c_\mathrm{p}$ is the prediction cost of source surrogate.
For most surrogates, the prediction cost is significantly lower than the training cost, enabling the proposed BCKT to enhance a wide range of SAS algorithms without introducing significant computational overhead.

%################################################################################
%######################################divider######################################
%################################################################################

\section{Experimental Results and Analyses}

\label{section:exp}

To empirically validate the effectiveness of the proposed MSAS-BCKT, we conduct a series of experiments in this section and thoroughly analyze the corresponding results.

\subsection{Experimental Settings}
In our experimental study, we considered two benchmark suites\footnote{Detailed information about these benchmarks is available in the supplement accompanying this paper.} to validate the proposed MSAS-BCKT.
The first suite consists of 9 single-objective multi-task optimization problems (MTOPs), with each MTOP containing two optimization tasks.
These problems are categorized into three groups based on the similarity between their component tasks: high similarity (HS), medium similarity (MS), and low similarity (LS).
The second suite comprises 6 single-objective many-task optimization problems (MaTOPs), each containing five optimization tasks with customized task similarities.
The diverse task similarities are considered to reflect the varying similarity relationships encountered in real-world scenarios, thereby facilitating a more comprehensive evaluation of different knowledge transfer algorithms~\cite{xue2023scalable,scott2024varying}.

\begin{figure}[ht]
	\centering
	\includegraphics[width=3.5in]{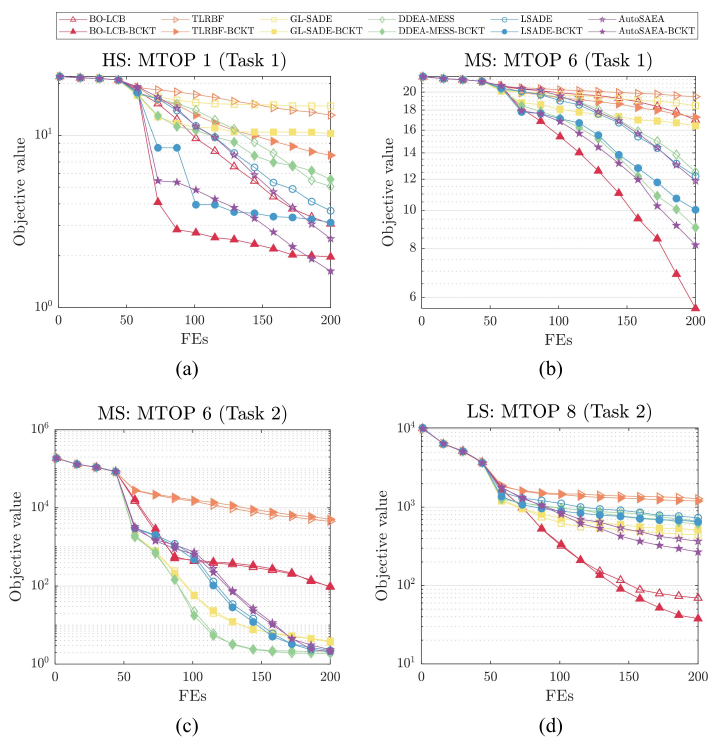}
	\caption{Averaged convergence curves of MSAS-BCKT and MSAS on three selected problems: (a) MTOP 1 (Task 1); (b) MTOP 6 (Task 1); (c) MTOP 6 (Task 2); (d) MTOP 8 (Task 2).}
	\label{fig:resultmulti}
\end{figure}

To examine the portability of the proposed BCKT, we employ six SAS algorithms to serve as backbone optimizers, including 1) the canonical Bayesian optimization algorithm using the lower confidence bound as its infill criterion (BO-LCB)~\cite{shahriari2016taking}; 2) the three-level radial basis function-assisted optimizer~\cite{li2022three}; 3) the global and local surrogate-assisted differential evolution~\cite{wang2023surrogate}; 4) the multi-evolutionary sampling strategy-based optimizer~\cite{yu2022data}; 5) the combined surrogate-based solver~\cite{jakub2023combining}; and 6) the surrogate-assisted optimization with model and infill criterion auto-configuration~\cite{xie2024surrogate} .
The parameter settings for these six optimizers align with those presented in~\cite{xue2024surrogate}.
In this paper, we refer to SOLVER-BCKT as the optimization algorithm `SOLVER' equipped with the proposed BCKT, while SOLVER denotes the original algorithm without knowledge transfer.
The general parameter settings are presented as follows:
\begin{enumerate}
    \item The number of initial solutions generated by the Latin hypercube sampling: 50.
    \item The maximum number of FEs for each MTOP: 200$\times$2.
    \item The maximum number of FEs for each MaTOP: 200$\times$5.
    \item The transfer interval in terms of real evaluation ($\delta$): 10.
    \item The prespecified uncertainty $\sigma_I^2$: $0.05^2$.
    \item The number of independent runs: 30.
\end{enumerate}

\subsection{Portability of BCKT}

To verify whether the proposed BCKT can enhance various SAS algorithms in a plug-and-play manner, we compare MSAS-BCKT with the base MSAS, utilizing six backbone optimizers across the nine two-task benchmark problems.
Specifically, we use the Wilcoxon rank-sum test with a significance level $\alpha=0.05$ to determine whether each MSAS-BCKT algorithm significantly outperforms its corresponding backbone MSAS based on their final optimization results.
For brevity, we denote the significant superiority, comparable performance, and significant inferiority of MSAS-BCKT compared to MSAS as win, tie, and loss, respectively.

\begin{figure}[ht]
	\centering
	\includegraphics[width=3.5in]{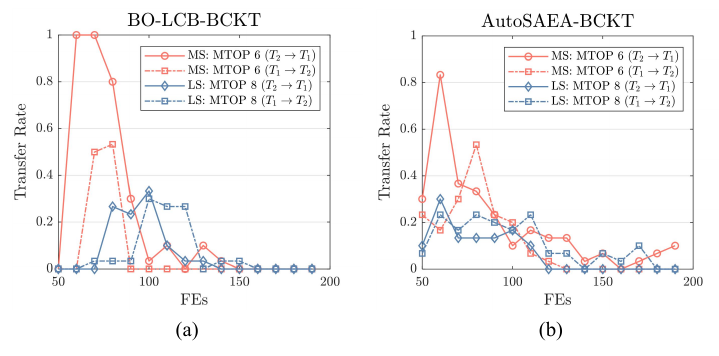}
	\caption{Averaged transfer rates of BO-LCB-BCKT and AutoSAEA-BCKT on MTOP 6 and MTOP 8: (a) BO-LCB-BCKT; (b) AutoSAEA-BCKT.}
	\label{fig:trans}
\end{figure}

Table \ref{tab:multi-table} summarizes the comparison results between MSAS-BCKT and MSAS with the 6 backbone optimizers on the 9 MTOPs with 18 individual tasks\footnote{The detailed optimization results are reported in the supplement document.}.
It is evident that the proposed BCKT achieves significant performance improvements across various backbone optimizers for both the HS and MS problems, highlighting its effectiveness in automatically identifying and leveraging promising cross-task solutions.
In the case of the LS problems, MSAS-BCKT demonstrates comparable optimization results to MSAS on the majority of the problems, highlighting the empirically nonnegative performance gain of BCKT for MSAS.
To elaborate further, Fig. \ref{fig:resultmulti} compares the averaged convergence curves of MSAS-BCKT and MSAS across the six optimizers on three selected problems, encompassing all three types of similarities.
The certain task similarities in terms of optimal solutions in the HS and MS problems enable the proposed BCKT to automatically identify promising cross-task solutions through its real-time transferability estimation, thereby leading to significant performance improvements, as can be observed from Fig. \ref{fig:resultmulti}(a) to Fig. \ref{fig:resultmulti}(b).
However, these performance improvements may be one-sided.
Comparing Fig. \ref{fig:resultmulti}(b) and Fig. \ref{fig:resultmulti}(c), we observe that task $T_1$ in MTOP 6 benefits from the optimization of task $T_2$, while $T_1$ offers no support to $T_2$.
On the LS problems, many existing multi-task surrogate-assisted optimization algorithms experience performance slowdowns due to the threat of negative transfer, as will be demonstrated later.
Promisingly, as shown in Fig. \ref{fig:resultmulti}(d), the proposed MSAS-BCKT can consistently mitigate the risk of negative transfer by effectively managing knowledge transfer through the Bayesian transferability estimation.
This advantage is particularly significant when a priori transferability criteria inaccurately indicate a high likelihood of knowledge transfer for tasks with intrinsically low transferability.

To examine the intermediate decisions on knowledge transfer in MSAS-BCKT, we present the averaged transfer rates of BO-LCB-BCKT and AutoSAEA-BCKT over 30 runs on MTOP 6 and MTOP 8 in Fig. \ref{fig:trans}.
During the optimization process, the transfer rate from task $T_i$ to task $T_j$ will be recorded as one when knowledge transfer occurs from $T_i$ to $T_j$, and zero otherwise.
As shown by the results of BO-LCB-BCKT in Fig. \ref{fig:trans}(a), the proposed BCKT demonstrates distinct transfer rate patterns for problems with different types of similarities.
\begin{table}[ht]
	\caption{Wins, ties and losses of MSAS-BCKT against MSAS with the 6 backbone optimizers on the 6 MaTOPs.}
	\centering
	\footnotesize
	\heavyrulewidth=0.12em
	\lightrulewidth=0.1em
	\cmidrulewidth=0.1em
	\setlength\tabcolsep{3pt}
	\begin{tabular}{lcccc}
		\toprule
Optimizer&HS Problems&MS Problems&LS Problems&Summary\\
		\midrule
		BO-LCB~\cite{shahriari2016taking}&4/6/0&1/8/1&0/10/0&5/24/1\\
		\midrule
		TLRBF~\cite{li2022three}&5/5/0&1/9/0&0/10/0&6/24/0\\
		\midrule
		GL-SADE~\cite{wang2023surrogate}&4/6/0&2/8/0&0/10/0&6/24/0\\
		\midrule
		DDEA-MESS~\cite{yu2022data}&4/6/0&0/10/0&0/9/1&4/25/1\\
		\midrule
		LSADE~\cite{jakub2023combining}&5/5/0&0/10/0&2/8/0&7/23/0\\
		\midrule
		AutoSAEA~\cite{xie2024surrogate}&4/6/0&1/9/0&0/9/1&5/24/1\\
		\midrule
		w/t/l &26/34/0&5/54/1&2/56/2&33/144/3\\
		\bottomrule
	\end{tabular}
	\label{tab:many-table}
\end{table}
Particularly, it can effectively capture the unsymmetrical transferability between $T_1$ and $T_2$ in MTOP 6, aligning well with the one-sided performance improvements shown in Fig. \ref{fig:resultmulti}(b) and Fig. \ref{fig:resultmulti}(c).
In addition, the Bayesian inference enhances the responsiveness of BCKT to the quality of both internal and external solutions, enabling adaptive use of external solutions through knowledge transfer to accelerate the target search in the early stages of optimization.
This is complemented by an automatic switchover to the standard SAS for further optimization, as shown in Fig. \ref{fig:trans}(a).
Regarding MTOP 8 with low similarity, the proposed BCKT can stably suppress the transferability to mitigate the risk of negative transfer, allowing for a comparable performance between BO-LCB-BCKT and BO-LCB, as demonstrated in Fig. \ref{fig:resultmulti}(d).
Similar observations can be obtained from the transfer rate results of AutoSAEA-BCKT shown in Fig. \ref{fig:trans}(b).
In summary, the intermediate decisions made by BCKT effectively elucidate its results, highlighting its interpretability and generalizability.

\subsection{Scalability of BCKT}

\begin{figure}[ht]
	\centering
	\includegraphics[width=3.5in]{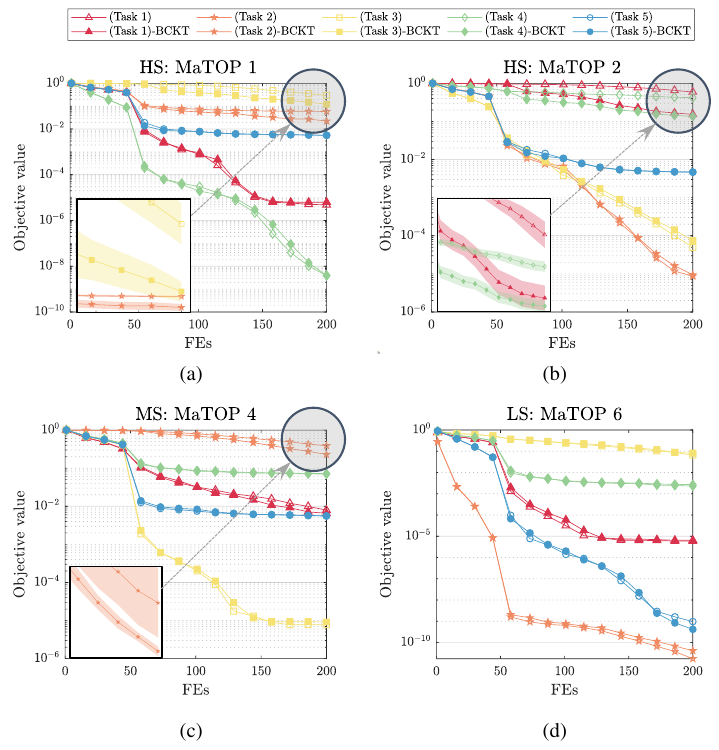}
	\caption{Averaged convergence curves of AutoSAEA-BCKT against AutoSAEA on four selected MaTOPs: (a) MaTOP 1; (b) MaTOP 2; (c) MaTOP 4; (d) MaTOP 6.}
	\label{fig:resultmany}
\end{figure}

The six MaTOPs are employed to validate the scalability of our proposed BCKT.
Each MaTOP consists of five individual tasks with grouped similarities, where inner-group tasks exhibit greater similarities than inter-group tasks.
Table \ref{tab:many-table} summarizes the comparison results between MSAS-BCKT and MSAS with the 6 backbone optimizers on the 6 MaTOPs.
It can be observed that the proposed BCKT exhibits effective performance improvements on the HS and MS problems while achieving comparable results with the baselines on the LS problems.
Fig. \ref{fig:resultmany} shows the averaged convergence curves of AutoSAEA-BCKT against AutoSAEA on four representative MaTOPs, wherein the final results, showing statistically significant differences, are presented with a zoomed-in view, including their 1/2 standard deviations.
More detailed optimization results obtained by the other optimizers, as well as detailed information about the six MaTOPs, can be found in the supplementary document accompanying this paper.
We observe that AutoSAEA-BCKT outperforms AutoSAEA on specific tasks within the HS and MS problems, while demonstrating comparable performance on MaTOP 6, which features low task similarities.
The results are consistent with those obtained on the MTOPs, highlighting BCKT's ability to harness the benefits of positive transfer while effectively mitigating the risks of negative transfer when addressing many-task EOPs.

\begin{figure}[ht]
	\centering
	\includegraphics[width=3.5in]{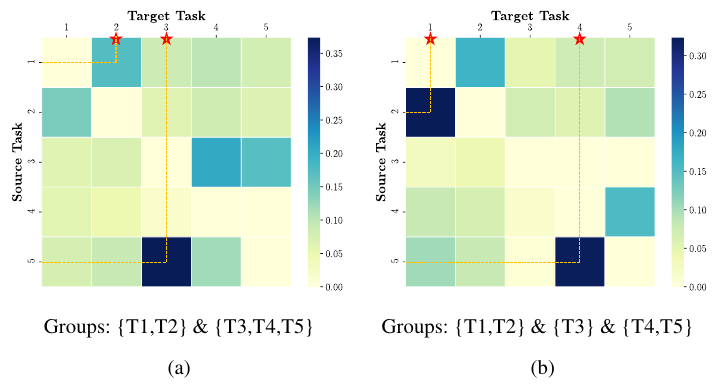}
	\caption{Averaged overall transfer rates of AutoSAEA-BCKT on MaTOP 1 and MaTOP 2: (a) MaTOP 1; (b) MaTOP 2.}
	\label{fig:hot}
\end{figure}

\begin{table}[ht]
	\caption{Wins, ties and losses of BO-LCB-BCKT against the six peer algorithms on the 9 MTOPs.}
	\centering
	\footnotesize
	\heavyrulewidth=0.12em
	\lightrulewidth=0.1em
	\cmidrulewidth=0.1em
	\setlength\tabcolsep{3pt}
	\begin{tabular}{lccccc}
		\toprule
Optimizer&HS Problems&MS Problems&LS Problems&Summary\\
		\midrule
		MaTDE~\cite{chen2020adaptive}&5/0/1&5/0/1&6/0/0&16/0/2\\
		\midrule
		MFEA-II~\cite{bali2020multifactorial}&6/0/0&5/0/1&6/0/0&17/0/1\\
		\midrule
		MMaTEA-DGT~\cite{lin2024multiobjective}&6/0/0&5/1/0&6/0/0&17/1/0\\
		\midrule
		LCB-EMT~\cite{wang2024evolutionary}&6/0/0&5/0/1&6/0/0&17/0/1\\
		\midrule
		RAMTEA~\cite{shen2024surrogate}&6/0/0&5/1/0&6/0/0&17/1/0\\
		\midrule
		SELF~\cite{tan2024surrogate}&3/1/2&5/0/1&6/0/0&14/1/3\\
		\midrule
		w/t/l &32/1/3&30/2/4&36/0/0&98/3/7\\
		\bottomrule
	\end{tabular}
	\label{tab:adaptivity}
\end{table}
\begin{figure}[ht]
	\centering
	\includegraphics[width=3.5in]{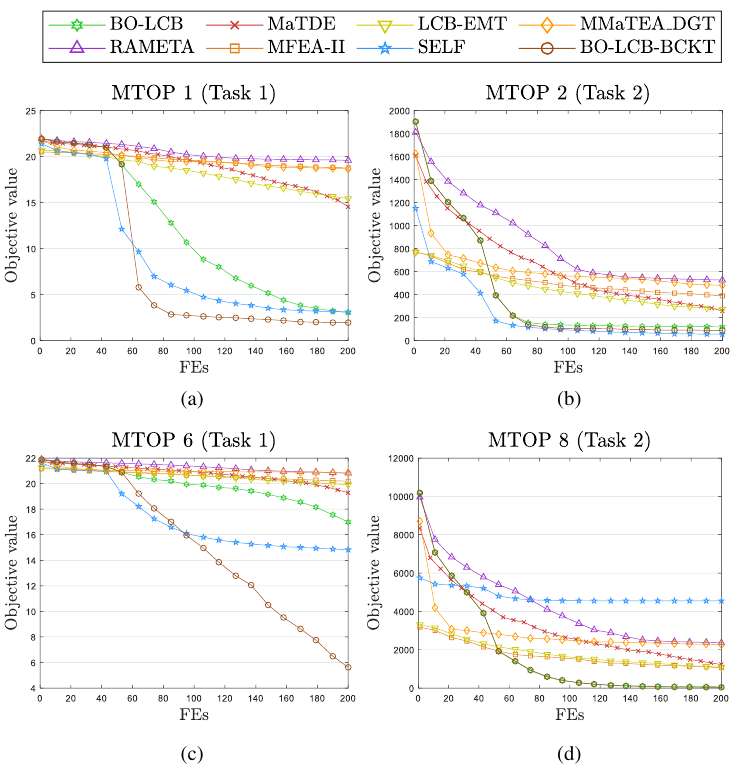}
	\caption{Averaged convergence curves of BO-LCB-BCKT and BO-LCB against the six peer algorithms on four representative optimization tasks: (a) MTOP 1 (Task 1); (b) MTOP 2 (Task 2); (c) MTOP 6 (Task 1); (d) MTOP 8 (Task 2).}
	\label{fig:comparison}
\end{figure}
To examine the intermediate decisions on knowledge transfer when solving MaTOPs, we present the averaged overall transfer rates of AutoSAEA-BCKT on MaTOP 1 and MaTOP 2 in Fig. \ref{fig:hot}.
The overall transfer rate is calculated by averaging the transfer rates recorded at different moments of knowledge competition.
In the heat maps presented in Fig. \ref{fig:hot}, the source tasks in a MaTOP are indexed by the rows, while the target tasks are indexed by the columns.
Each target task that achieved positive transfer is marked with a red star, and its source task with the highest transfer rate is highlighted with an orange dotted line.
The knowledge transfer decisions made by the proposed BCKT align closely with the grouped similarities illustrated at the bottom of Fig. \ref{fig:hot}, providing further insight into the corresponding performance improvements.
Concerning the third task $T_3$ of MaTOP 2, its low transfer rates associated with all the source tasks are intuitively graspable---it is isolated from them based on the grouped similarities.
In summary, the proposed MSAS-BCKT demonstrates scalable performance in addressing MaTOPs by automatically discovering and utilizing the underlying transferability of cross-task solutions.

\subsection{Adaptability of BCKT}

\begin{figure}[ht]
	\centering
	\includegraphics[width=3.5in]{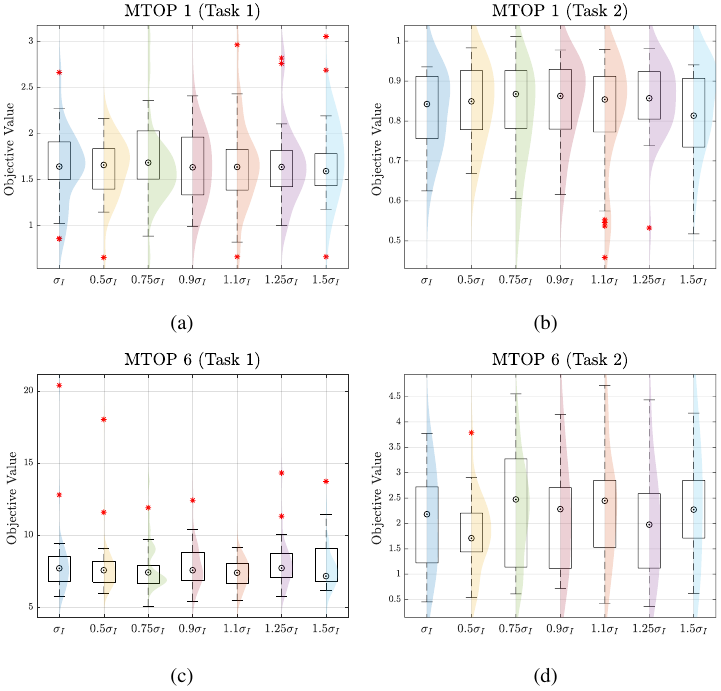}
	\caption{Final objective values obtained by AutoSAEA-BCKT with the seven configurations of $\sigma^2_I$ over 30 runs on two selected problems: (a) MTOP 1 (Task 1); (b) MTOP 1 (Task 2); (c) MTOP 6 (Task 1); (d) MTOP 6 (Task 2).}
	\label{fig:sensitivity}
\end{figure}

To evaluate the adaptability of the proposed MSAS-BCKT in promoting positive transfers while mitigating negative ones, we compare BO-LCB and BO-LCB-BCKT against six peer algorithms that concentrate on the issue of ``when to transfer'', including three advanced population-based algorithms, i.e., MaTDE~\cite{chen2020adaptive}, MFEA-II~\cite{bali2020multifactorial} and MMaTEA-DGT~\cite{lin2024multiobjective}, as well as three state-of-the-art model-based algorithms, namely LCB-EMT~\cite{wang2024evolutionary}, RAMTEA~\cite{shen2024surrogate} and SELF~\cite{tan2024surrogate}.
To ensure fair comparisons, the population-based algorithms are assigned the same number of real evaluations as the model-based algorithms, utilizing a population size of 10 over 20 generations.
Table \ref{tab:adaptivity} summarizes the wins, ties and losses of BO-LCB-BCKT against the six peer algorithms on the 9 MTOPs\footnote{More detailed optimization results are provided in the supplement.}.
The averaged convergence curves of BO-LCB and BO-LCB-BCKT, compared to the six peer algorithms across four representative optimization tasks with distinct types of task similarities, are presented in Fig. \ref{fig:comparison}.
It can be observed that the proposed BO-LCB-BCKT can achieve significantly better optimization results than the remaining algorithms.
Particularly, as shown in Fig. \ref{fig:comparison}(a) and Fig. \ref{fig:comparison}(c), our BO-LCB-BCKT and the advanced optimizer SELF demonstrate effective convergence speedups compared to the baseline solver BO-LCB, highlighting their ability to identify and leverage underlying inter-task synergies for improved optimization performance.
However, such convergence speedups are not guaranteed, even with certain task similarities, as illustrated by the results on the second task of MTOP 2 in Fig. \ref{fig:comparison}(b), wherein both BO-LCB-BCKT and SELF exhibit comparable optimization to BO-LCB.
Additionally, we can observe a performance slowdown of SELF compared to BO-LCB on the second task of MTOP 8, attributable to its low task similarity, demonstrating the inability of SELF to avoid negative transfer.
In contrast, by effectively suppressing the transfer of conflicting cross-task solutions, our proposed BO-LCB-BCKT can achieve comparable optimization performance to BO-LCB when tackling such MTOP with low task similarities.
In summary, the proposed MSAS-BCKT algorithm outperforms the six state-of-the-art algorithms in addressing multi-task EOPs, primarily due to the portability of BCKT, which ensures its compatibility with various advanced SAS backbones, and its adaptability in fostering positive transfers while mitigating negative ones.

\begin{figure}[ht]
	\centering
	\includegraphics[width=3.5in]{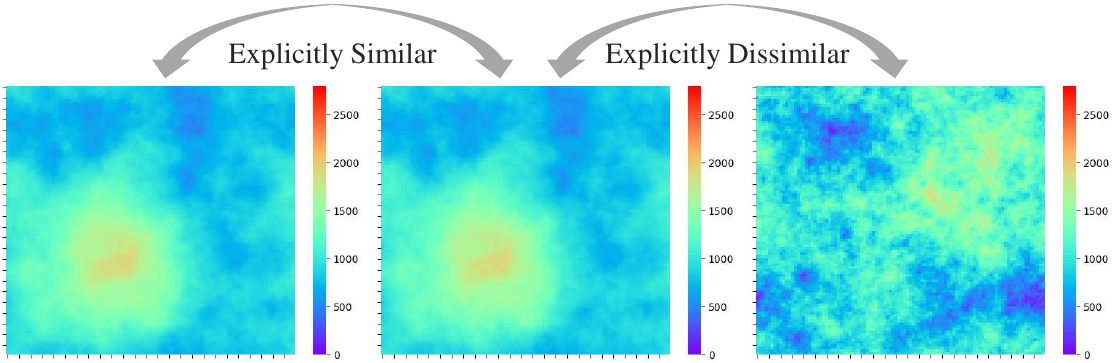}
	\caption{Permeability fields of the three reservoir models.}
	\label{fig:permx}
\end{figure}

\subsection{Sensitivity Analysis}

There are two prespecified parameters that may influence the performance of the proposed MSAS-BCKT, including the transfer interval $\delta$ and the uncertainty of transferability assessment $\sigma_I$.
Regarding the transfer interval $\delta$, the insensitivity of CKT to this parameter has been empirically validated in~\cite{xue2024surrogate}.
The study found that varying configurations of $\delta$ do not affect the decisions on knowledge transfer during the optimization process, which effectively explains the consistent performance improvements achieved by CKT.
Concerning the uncertainty parameter $\sigma_I$, we employ AutoSAEA as the backbone optimizer and vary the prespecified uncertainty value $\sigma_I$ in BCKT to generate six additional configurations, including $\pm0.1\sigma_I$, $\pm0.25\sigma_I$ and $\pm0.5\sigma_I$.
The final objective values obtained by AutoSAEA-BCKT with the seven configurations of $\sigma_I$ over 30 runs on MTOP 1 and MTOP 6 are presented as boxplots in Fig. \ref{fig:sensitivity}.
For better illustration, the estimated density of the objective values obtained by each configuration of $\sigma_I$ is included alongside its corresponding boxplot.
We can observe that the AutoSAEA-BCKT algorithms, equipped with different values of $\sigma_I$, produce comparable final objective values on both MTOP 1 and MTOP 6, demonstrating the insensitivity of BCKT to $\sigma_I$.
However, despite this insensitivity, a very large value of $\sigma_I$ is not recommended, as it can lead to excessive divergence in initial transferability and heighten the risk of negative transfer.
This underscores the importance of accurately quantifying the uncertainty and adopting transferability assessment methods with lower uncertainties whenever possible, aligning with the implications of Eq. \eqref{eq:fisher_general}.

\subsection{A Practical Case Study: Well Placement Optimization}

Well placement optimization in the field of petroleum engineering is typically computationally expensive due to its reliance on high-fidelity simulations.
Whenever we encounter a new optimization task, it typically requires substantial computational resources to achieve satisfactory results when starting from scratch.
Fortunately, when faced with multiple well placement optimization tasks, we have the opportunity to leverage their underlying synergies for enhanced multitasking optimization performance.
\begin{figure}[ht]
	\centering
	\includegraphics[width=2.4in]{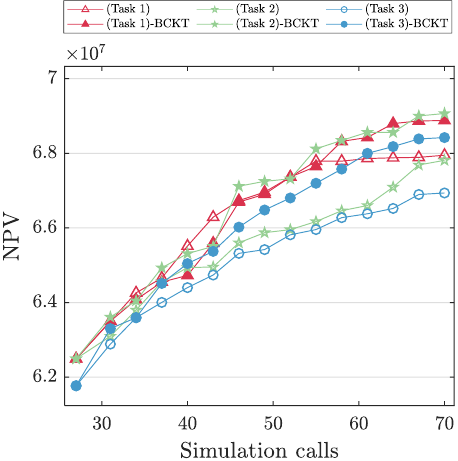}
	\caption{Averaged NPV curves of BO-LCB and BO-LCB-BCKT on the three well placement optimization tasks.}
	\label{fig:oil1}
\end{figure}
Therefore, we employ an expensive MaTOP with three individual well placement optimization tasks to examine the effectiveness of the proposed MSAS-BCKT.
In particular, the net present value (NPV) is used as the objective function to measure the economic benefits of specific well placements, which is given by
\begin{equation}
\small
NPV\left(\boldsymbol{v}\right) = \sum_{t=1}^{T} \frac{r_o Q^t_{po}\left(\boldsymbol{v}\right) - r_{wd} Q^t_{pw}\left(\boldsymbol{v}\right) - r_{wi} Q^t_{iw}\left(\boldsymbol{v}\right)}{(1 + r)^t},
\label{eq:exp}
\end{equation}
where $NPV\left(\boldsymbol{v}\right)$ denotes the NPV value of the well placement scheme $\boldsymbol{v}$, $t$ is the $t$-th timestep, $T$ denotes the total number of timesteps, $Q^t_{po}$ and $Q^t_{pw}$ denote the oil and water outputs of all the production wells at time $t$, $Q_{iw}^t$ represents the water input of all the injection wells at time $t$, $r_o$ denotes the oil revenue, $r_{wd}$ is the the cost of disposing of produced water, $r_{wi}$ denotes the cost of water injection, $r$ denotes the discount rate.
The calculation of the NPV for a single candidate solution of well placement scheme, as shown in Eq. \eqref{eq:exp}, typically relies on high-fidelity subsurface flow simulations.
While these simulations provide accurate results, they are computationally expensive, often taking from several minutes to a few days to complete.
As a result, the total number of NPV calls (i.e., FEs) is significantly constrained in practical applications.

The permeability fields of the three reservoir models associated with the three well placement optimization tasks are illustrated in Fig. \ref{fig:permx}.
The first two reservoirs have similar permeability fields, while the third reservoir's permeability field differs significantly from the first two.
This variation is intended to mimic the varying task similarities observed in real-world MaTOPs.
Each reservoir model will be developed using the nine-spot well pattern, featuring five production wells and four injection wells, with a simulation grid size of $100\times100\times1$.
For brevity, the remaining reservoir properties are not presented here but can be found in~\cite{chen2020global}.
The maximum number of simulation calls for optimizing each individual well placement optimization task is 70.

Without loss of generality, we employ the popular BO-LCB optimizer as the backbone to validate the applicability of the proposed BCKT to the multi-task well placement optimization.
Fig. \ref{fig:oil1} compares the averaged convergence curves of BO-LCB and BO-LCB-BCKT on the three well placement optimization tasks.
It can be seen that the proposed BCKT can effectively improve the search performance of BO-LCB across the three tasks.
On the one hand, the convergence speed and the final solution quality of the second task is significantly enhanced by the proposed BCKT, attributable to its explicit similarity to the first task that facilitates positive transfer of cross-task solutions.
On the other hand, the search performance of BO-LCB on the third task can be effectively improved despite its explicit dissimilarity to the first two tasks, demonstrating the proposed BCKT's capability to identify and leverage latent synergies.
These observations align well with the benchmark results, demonstrating the adaptability of BCKT in promoting positive transfer while suppressing negative transfer when addressing practical multi-task EOPs.

%################################################################################
%######################################divider######################################
%################################################################################

\section{Conclusions}

\label{section:con}

To alleviate the cold-start issue when solving EOPs with SAS, this paper has introduced a novel Bayesian competitive knowledge transfer (BCKT) method that exploits the inter-task synergies of multiple EOPs in a multitasking environment, thereby facilitating enhanced search performance.
Apart from its superior portability and scalability, which make it effective in enhancing various SAS algorithms for solving both multi-task and many-task EOPs, the proposed BCKT method distinguishes itself from peer algorithms through its outstanding adaptability in knowledge transfer.
Specifically, by estimating the transferability of source elite solutions from a Bayesian perspective that accommodates both prior beliefs and empirical evidence, we can initiate a competition between these source elite solutions and the target promising solution to identify the winning candidate for real evaluation, thereby facilitating adaptive knowledge transfer across different EOPs.
Moreover, the flexibility to incorporate prior beliefs or criteria allows our method to seamlessly integrate diverse sources of transferability information, leading to more accurate adaptive knowledge transfer in complex scenarios.
In addition to confirming the asymptotic unbiasedness and efficiency of our transferability estimation, we have conducted extensive empirical studies to demonstrate the effectiveness of BCKT in solving both multi-task and many-task EOPs.
Furthermore, the superiority of BCKT over state-of-the-art methods, along with its applicability to a practical case study in the petroleum industry, has been empirically validated.

In future work, we will expand the concept of integrating both prior beliefs and empirical evidence in BCKT to tackle other key issues in knowledge transfer, such as source selection in ``what to transfer'' and the mapping of learning in ``how to transfer''.
Additionally, we are interested in generalizing ``promising'' knowledge for transfer in external improvement and ``positive'' transfer for performance evaluation, incorporating more sophisticated infill criteria.
Furthermore, we intend to conduct in-depth investigations into the precise quantification of Fisher information within the context of evolutionary transfer optimization, which will deepen our understanding of the key factors that contribute to effective and generalizable knowledge transfer. 
Finally, we are eager to explore the generalization of Bayesian transferability estimation to address multiobjective and constrained optimization tasks.
%\section*{Acknowledgment}

%This work is supported ...

\footnotesize
\bibliography{MyBiBfile}

\newpage
~
\newpage

\normalsize

\section*{Supplementary Document}

\setcounter{figure}{0}
\setcounter{table}{0}
\renewcommand\thesection{S-\Roman{section}}
\renewcommand\thefigure{S-\arabic{figure}}
\renewcommand\thedefi{S-\arabic{defi}}
\renewcommand\thetable{S-\arabic{table}}

\subsection{Multi-Task and Many-Task Instances}

Table \ref{tab:multi-problem} and Table \ref{tab:many-problem} present the detailed information about the multi-task and many-task benchmark suites used in this study, wherein each multi-task optimization problem (MTOP) contains two tasks while each many-task optimization problem (MaTOP) possesses five tasks.
For each MTOP, two individual tasks are treated as belonging to a single group, with adjustable task similarity.
First, the group center is generated randomly to avoid the trivial optimum consisting entirely of zeros, as given by $\boldsymbol{o}_g\in\lbrack0,1\rbrack^d$ drawn from the uniform distribution.
Subsequently, the optimal solution of the $i$-th task in the common search space, as denoted by $\boldsymbol{o}^i$, is configured as follows:
\begin{equation}
\boldsymbol{o}^i = \boldsymbol{o}_g+\beta*\boldsymbol{r},
\label{eq:group_center}
\end{equation}
where $\beta$ represents a perturbation coefficient for adjusting the task similarity of the group, $\boldsymbol{r}$ denotes a random vector within $\lbrack-1,1\rbrack^d$.
The elements of $\boldsymbol{o}^i$ beyond the boundary will be truncated.
In this study, we consider three cases of $\beta$, i.e., $\beta_\mathrm{h}=0$, $\beta_\mathrm{m}=0.2$ and $\beta_\mathrm{l}=0.6$, to represent three classes of MTOPs with high, medium and low similarity, respectively, as shown in Table \ref{tab:multi-problem}.
Lastly, the optimum solution of the $i$-th task, as denoted by $\boldsymbol{x}^i_*$, is determined as follows:
\begin{equation}
\boldsymbol{x}^i_*=\boldsymbol{x}^i_{\mathrm{lb}}+\left(\boldsymbol{x}^i_{\mathrm{ub}}-\boldsymbol{x}^i_{\mathrm{lb}}\right)\times\boldsymbol{o}^i,
\label{eq:shift}
\end{equation}
where $\boldsymbol{x}^i_{\mathrm{lb}}$ and $\boldsymbol{x}^i_{\mathrm{ub}}$ represent the lower and upper bounds of the $i$-th task's search space.

\begin{table}[ht]
	\caption{Detailed information about the nine MTOPs.}
	\centering
	\renewcommand{\arraystretch}{1}
	\heavyrulewidth=0.12em
	\lightrulewidth=0.08em
	\cmidrulewidth=0.05em
	%\fontsize{5.5pt}{5.5pt}\selectfont
	\scriptsize
	\setlength\tabcolsep{3.5pt}
	\resizebox{0.5\textwidth}{!}{
\begin{tabular}{*{6}{lccccc}}
	\toprule
	\makecell[c]{Test\\Problem}&\makecell[c]{Task\\ID}&\makecell[c]{Objective\\Function}&\makecell[c]{Search\\Space}&\makecell[c]{Task\\Similarity}\\
	\midrule
	\multirow{2.5}*{MTOP 1}&\multirow{1}*{$T_1$}&Ackley&$[-32, 32]^{10}$&\multirow{10}*{\makecell[c]{High Similarity \\ (HS)\\$\beta_\mathrm{h}=0$}}\\
	\cmidrule(){2-4}
	&\multirow{1}*{$T_2$}&Griewank&$[-200, 200]^{10}$\\
	\cmidrule(){1-4}
	\multirow{2.5}*{MTOP 2}&\multirow{1}*{$T_1$}&Sphere&$[-100, 100]^{15}$\\
	\cmidrule(){2-4}
	&\multirow{1}*{$T_2$}&Rastrigin&$[-10, 10]^{15}$\\
	\cmidrule(){1-4}
	\multirow{2.5}*{MTOP 3}&\multirow{1}*{$T_1$}&Elliptic&$[-50, 50]^{20}$\\
	\cmidrule(){2-4}
	&\multirow{1}*{$T_2$}&Rosenbrock&$[-50, 50]^{20}$\\
	\midrule
	\multirow{2.5}*{MTOP 4}&\multirow{1}*{$T_1$}&Griewank&$[-200, 200]^{20}$&\multirow{10}*{\makecell[c]{Medium Similarity \\ (MS)\\$\beta_\mathrm{m}=0.2$}}\\
	\cmidrule(){2-4}
	&\multirow{1}*{$T_2$}&Weierstrass&$[-0.5, 0.5]^{20}$\\
	\cmidrule(){1-4}
	\multirow{2.5}*{MTOP 5}&\multirow{1}*{$T_1$}&Schwefel&$[-10, 10]^{10}$\\
	\cmidrule(){2-4}
	&\multirow{1}*{$T_2$}&Levy&$[-30, 30]^{10}$\\
	\cmidrule(){1-4}
	\multirow{2.5}*{MTOP 6}&\multirow{1}*{$T_1$}&Ackley&$[-32, 32]^{20}$\\
	\cmidrule(){2-4}
	&\multirow{1}*{$T_2$}&Sphere&$[-100, 100]^{20}$\\
	\midrule
	\multirow{2.5}*{MTOP 7}&\multirow{1}*{$T_1$}&Sphere&$[-50, 50]^{15}$&\multirow{10}*{\makecell[c]{Low Similarity \\ (LS)\\$\beta_\mathrm{l}=0.6$}}\\
	\cmidrule(){2-4}
	&\multirow{1}*{$T_2$}&Rastrigin&$[-10, 10]^{15}$\\
	\cmidrule(){1-4}
	\multirow{2.5}*{MTOP 8}&\multirow{1}*{$T_1$}&Quartic&$[-5, 5]^{20}$\\
	\cmidrule(){2-4}
	&\multirow{1}*{$T_2$}&Levy&$[-30, 30]^{20}$\\
	\cmidrule(){1-4}
	\multirow{2.5}*{MTOP 9}&\multirow{1}*{$T_1$}&Ackley&$[-32, 32]^{10}$\\
	\cmidrule(){2-4}
	&\multirow{1}*{$T_2$}&Griewank&$[-200, 200]^{10}$\\
	\bottomrule
	\end{tabular}
	}
	\label{tab:multi-problem}
\end{table}

For each MaTOP, we propose simulating the modular nature of task groups by configuring the locations of group centers independently, to mimic scenarios where different task groups originate from distinct domains.
Suppose there are $m$ groups in an MaTOP, the locations of their centers are configured independently, as denoted by $\boldsymbol{o}^j_g\in\lbrack0,1\rbrack^d,\,\,1\le j\le m$.
Next, the optimal solution for each task within a specific group is determined using Eqs. \eqref{eq:group_center} and \eqref{eq:shift}.
The group information for the six MaTOPs is provided in the second column of Table \ref{tab:many-problem}.
We also examine three types of task similarities for these MaTOPs, defined using different perturbation coefficients, as shown in the last column of Table \ref{tab:many-problem}.

\begin{table}[ht]
	\caption{Detailed information about the six MaTOPs.}
	\centering
	\renewcommand{\arraystretch}{1}
	\heavyrulewidth=0.12em
	\lightrulewidth=0.08em
	\cmidrulewidth=0.05em
	%\fontsize{5.5pt}{5.5pt}\selectfont
	\scriptsize
	\setlength\tabcolsep{3.5pt}
	\resizebox{0.5\textwidth}{!}{
\begin{tabular}{*{6}{lccccc}}
	\toprule
	\makecell[c]{Test\\Problem}&\makecell[c]{Task\\Group}&\makecell[c]{Task\\ID}&\makecell[c]{Objective\\Function}&\makecell[c]{Search\\Space}&\makecell[c]{Task\\Similarity}\\
	\midrule
	\multirow{8}*{MaTOP 1}&\multirow{3}*{$G_1$}&\multirow{1}*{$T_1$}&Sphere&$[-100, 100]^{15}$&\multirow{16}*{\makecell[c]{High\\Similarity \\ (HS)\\$\beta_\mathrm{h}=0$}}\\
	\cmidrule(){3-5}
	&&\multirow{1}*{$T_2$}&Rastrigin&$[-10, 10]^{15}$\\
	\cmidrule(){2-5}
	&\multirow{4.5}*{$G_2$}&\multirow{1}*{$T_3$}&Ackley&$[-32, 32]^{15}$\\
	\cmidrule(){3-5}
	&&\multirow{1}*{$T_4$}&Elliptic&$[-50, 50]^{15}$\\
	\cmidrule(){3-5}
	&&\multirow{1}*{$T_5$}&Griewank&$[-200, 200]^{15}$\\
	\cmidrule(){1-5}
	\multirow{8}*{MaTOP 2}&\multirow{3}*{$G_1$}&\multirow{1}*{$T_1$}&Ackley&$[-32, 32]^{20}$\\
	\cmidrule(){3-5}
	&&\multirow{1}*{$T_2$}&Sphere&$[-100, 100]^{20}$\\
	\cmidrule(){2-5}
	&\multirow{1}*{$G_2$}&\multirow{1}*{$T_3$}&Rosenbrock&$[-50, 50]^{20}$\\
	\cmidrule(){2-5}
	&\multirow{3}*{$G_3$}&\multirow{1}*{$T_4$}&Weierstrass&$[-0.5, 0.5]^{20}$\\
	\cmidrule(){3-5}
	&&\multirow{1}*{$T_5$}&Griewank&$[-200, 200]^{20}$\\
	\cmidrule(){1-6}
	\multirow{8}*{MaTOP 3}&\multirow{3}*{$G_1$}&\multirow{1}*{$T_1$}&Sphere&$[-100, 100]^{10}$&\multirow{16}*{\makecell[c]{Medium\\Similarity \\ (MS)\\$\beta_\mathrm{m}=0.2$}}\\
	\cmidrule(){3-5}
	&&\multirow{1}*{$T_2$}&Rastrigin&$[-10, 10]^{10}$\\
	\cmidrule(){2-5}
	&\multirow{3}*{$G_2$}&\multirow{1}*{$T_3$}&Elliptic&$[-50, 50]^{10}$\\
	\cmidrule(){3-5}
	&&\multirow{1}*{$T_4$}&Weierstrass&$[-0.5, 0.5]^{10}$\\
	\cmidrule(){2-5}
	&\multirow{1}*{$G_3$}&\multirow{1}*{$T_5$}&Schwefel&$[-10, 10]^{10}$\\
	\cmidrule(){1-5}
	\multirow{8}*{MaTOP 4}&\multirow{1}*{$G_1$}&\multirow{1}*{$T_1$}&Levy&$[-30, 30]^{15}$\\
	\cmidrule(){2-5}
	&\multirow{3}*{$G_2$}&\multirow{1}*{$T_2$}&Ackley&$[-32, 32]^{15}$\\
	\cmidrule(){3-5}
	&&\multirow{1}*{$T_5$}&Griewank&$[-200, 200]^{15}$\\
	\cmidrule(){2-5}
	&\multirow{3}*{$G_3$}&\multirow{1}*{$T_3$}&Sphere&$[-100, 100]^{15}$\\
	\cmidrule(){3-5}
	&&\multirow{1}*{$T_4$}&Rastrigin&$[-10, 10]^{15}$\\
	\cmidrule(){1-6}
	\multirow{8}*{MaTOP 5}&\multirow{1}*{$G_1$}&\multirow{1}*{$T_1$}&Schwefel&$[-10, 10]^{20}$&\multirow{16}*{\makecell[c]{Low\\Similarity \\ (LS)\\$\beta_\mathrm{l}=0.6$}}\\
	\cmidrule(){2-5}
	&\multirow{1}*{$G_2$}&\multirow{1}*{$T_2$}&Levy&$[-30, 30]^{20}$\\
	\cmidrule(){2-5}
	&\multirow{1}*{$G_3$}&\multirow{1}*{$T_3$}&Quartic&$[-5, 5]^{20}$\\
	\cmidrule(){2-5}
	&\multirow{1}*{$G_4$}&\multirow{1}*{$T_4$}&Rastrigin&$[-10, 10]^{20}$\\
	\cmidrule(){2-5}
	&\multirow{1}*{$G_5$}&\multirow{1}*{$T_5$}&Ackley&$[-32, 32]^{20}$\\
	\cmidrule(){1-5}
	\multirow{8}*{MaTOP 6}&\multirow{1}*{$G_1$}&\multirow{1}*{$T_1$}&Sphere&$[-100, 100]^{10}$\\
	\cmidrule(){2-5}
	&\multirow{1}*{$G_2$}&\multirow{1}*{$T_2$}&Schwefel&$[-10, 10]^{10}$\\
	\cmidrule(){2-5}
	&\multirow{1}*{$G_3$}&\multirow{1}*{$T_3$}&Weierstrass&$[-0.5, 0.5]^{10}$\\
	\cmidrule(){2-5}
	&\multirow{1}*{$G_4$}&\multirow{1}*{$T_4$}&Griewank&$[-200, 200]^{10}$\\
	\cmidrule(){2-5}
	&\multirow{1}*{$G_5$}&\multirow{1}*{$T_5$}&Elliptic&$[-50, 50]^{10}$\\
	\bottomrule
	\end{tabular}
	}
	\label{tab:many-problem}
\end{table}

%################################################################################
%######################################divider######################################
%################################################################################

\subsection{Results with Reduced Function Evaluations}

\begin{figure}[ht]
	\centering
	\includegraphics[width=3.5in]{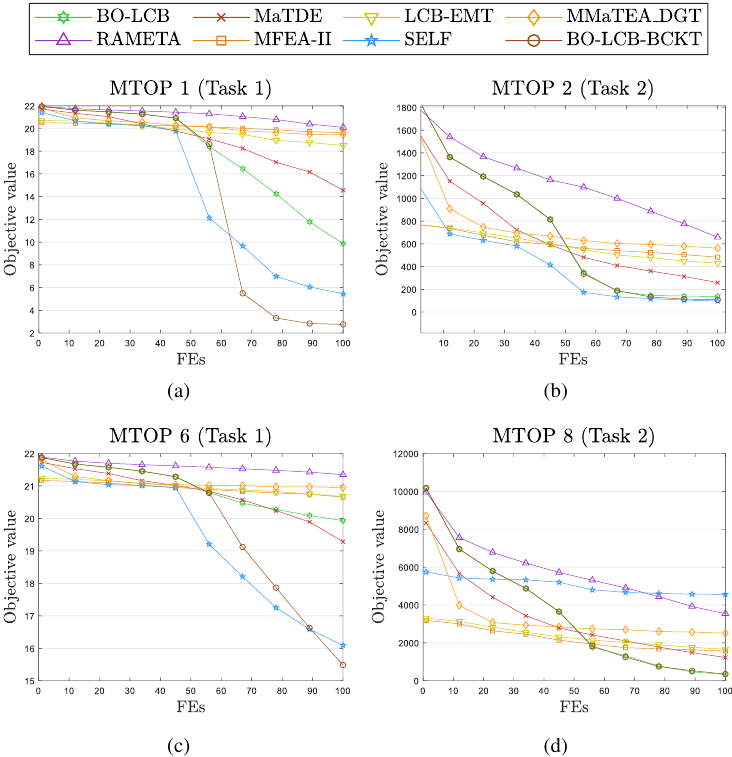}
	\caption{Averaged convergence curves of BO-LCB-BCKT and BO-LCB against six peer algorithms with 100 FEs per task on four representative optimization tasks: (a) MTOP 1 (Task 1); (b) MTOP 2 (Task 2); (c) MTOP 6 (Task 1); (d) MTOP 8 (Task 2).}
	\label{fig:comparison1}
\end{figure}

To evaluate the performance of the proposed MSAS-BCKT under limited computational budgets, we compared BO-LCB and BO-LCB-BCKT with six peer algorithms on 9 MTOPs with a reduced number of function evaluations.
The remaining experimental details are consistent with Section IV-D of the manuscript.
Table \ref{tab:adaptivity1} summarizes the wins, ties, and losses of BO-LCB-BCKT against six peer algorithms on nine MTOPs with a strict budget of 100 FEs per task.
Fig. \ref{fig:comparison1} presents the averaged convergence curves of BO-LCB and BO-LCB-BCKT in comparison with six baseline algorithms across four representative optimization tasks with 100 FEs per task.
As shown in Fig. \ref{fig:comparison1} and Table \ref{tab:adaptivity1}, the proposed MSAS-BCKT demonstrates a statistically significant advantage over six peer algorithms under a limited computational budget of 100 FEs per task, consistent with the experimental results reported in Section IV-D of the manuscript.
In conclusion, the proposed MSAS-BCKT algorithm exhibits strong applicability in resource-constrained environments.

%################################################################################
%######################################divider######################################
%################################################################################

\subsection{Detailed Optimization Results}

Tables \ref{tab:multi-result1} and \ref{tab:multi-result2} present the final results of MSAS-BCKT and MSAS equipped with the six backbone optimizers across the 9 MTOPs over 30 independent runs on the first and second component tasks, i.e., $T_1$ and $T_2$, respectively.
The best performing algorithms are highlighted based on the Wilcoxon rank-sum test with Holm $p$-value correction ($\alpha=0.05$).

Tables \ref{tab:many-result1}, Tables \ref{tab:many-result2}, Tables \ref{tab:many-result3}, Tables \ref{tab:many-result4} and \ref{tab:many-result5} respectively present the final results of MSAS-BCKT and MSAS equipped with the six optimizers across the 6 MaTOPs over 30 independent runs on the five component tasks.
The best performing algorithms are highlighted based on the Wilcoxon rank-sum test with Holm $p$-value correction ($\alpha=0.05$).

Tables \ref{tab:compare} presents the averaged objective values and their standard deviation obtained by MaTDE, MFEA-II, MMaTEA-DGT, LCB-EMT, RAMTEA, SELF and BO-LCB-BCKT on the 9 MTOPs across 30 independent runs.
The highlighted entries are significantly better based on the Wilcoxon rank-sum test with Holm $p$-value correction ($\alpha=0.05$).

\begin{table}[H]
	\caption{Wins, ties and losses of BO-LCB-BCKT against the six peer algorithms on the 9 MTOPs with 100 FEs per task.}
	\centering
	\footnotesize
	\heavyrulewidth=0.12em
	\lightrulewidth=0.1em
	\cmidrulewidth=0.1em
	\setlength\tabcolsep{3pt}
	\begin{tabular}{lccccc}
		\toprule
Optimizer&HS Problems&MS Problems&LS Problems&Summary\\
		\midrule
		MaTDE&5/0/1&5/0/1&6/0/0&16/0/2\\
		\midrule
		MFEA-II&6/0/0&5/0/1&6/0/0&17/0/1\\
		\midrule
		MMaTEA-DGT&6/0/0&5/1/0&6/0/0&17/1/0\\
		\midrule
		LCB-EMT&6/0/0&5/0/1&6/0/0&17/0/1\\
		\midrule
		RAMTEA&6/0/0&5/1/0&6/0/0&17/1/0\\
		\midrule
		SELF&3/1/2&5/0/1&6/0/0&14/1/3\\
		\midrule
		w/t/l &32/1/3&30/2/4&36/0/0&98/3/7\\
		\bottomrule
	\end{tabular}
	\label{tab:adaptivity1}
\end{table}

\begin{table*}[ht]
	\caption{Optimization results of MSAS-BCKT against MSAS with the six backbone optimizers on $T_1$ of the 9 MTOPs. Mean and standard deviation (i.e., mean$\pm$std) are based on 30 independent runs. The winner in each comparison between MSAS and MSAS-BCKT is highlighted based on the Wilcoxon rank-sum test with Holm $p$-value correction ($\alpha=0.05$).}
	\centering
	\heavyrulewidth=0.12em
	\lightrulewidth=0.08em
	\cmidrulewidth=0.05em
	%\fontsize{5.5pt}{5.5pt}\selectfont
	\scriptsize
	\setlength\tabcolsep{4.4pt}
\begin{tabular}{*{12}{lcccccccccccc}}
	\toprule
	\multirow{3}*{Optimizer}&\multirow{3}*{Transfer}&\multirow{3}*{Stat.}&\multicolumn{3}{c}{HS Problems}&\multicolumn{3}{c}{MS Problems}&\multicolumn{3}{c}{LS Problems}\\
	\cmidrule(l){4-6}\cmidrule(l){7-9}\cmidrule(l){10-12}
	&&&MTOP 1&MTOP 2&MTOP 3&MTOP 4&MTOP 5&MTOP 6&MTOP 7&MTOP 8&MTOP 9\\
	\midrule
	\multirow{6}*{BO-LCB}&\multirow{3}*{/}&mean&3.06e+00&2.90e+01&1.09e+08&1.15e+00&1.47e+02&1.70e+01&2.43e+01&3.09e+02&2.69e+00\\
	\cmidrule(){3-12}	&&std&1.95e+00&2.60e+01&3.31e+07&4.16e-02&2.51e+02&4.38e+00&1.20e+01&2.16e+02&4.75e-01\\
	\cmidrule(){2-12}
	&\multirow{3}*{BCKT}&mean&\cellcolor[rgb]{0.74,0.74,0.74}1.97e+00&3.38e+01&9.44e+07&1.14e+00&6.52e+01&\cellcolor[rgb]{0.74,0.74,0.74}5.63e+00&2.16e+01&2.65e+02&3.69e+00\\
	\cmidrule(){3-12}	&&std&4.00e-01&3.19e+01&3.45e+07&2.86e-02&4.18e+01&1.83e+00&6.31e+00&1.39e+02&3.33e+00\\
	\midrule
	\multirow{6}*{TLRBF}&\multirow{3}*{/}&mean&1.31e+01&1.92e+03&4.01e+07&6.71e+00&4.09e+01&1.95e+01&6.03e+02&5.39e+02&1.32e+01\\
	\cmidrule(){3-12}	&&std&3.65e+00&1.21e+03&1.92e+07&1.58e+00&2.66e+01&1.21e+00&3.74e+02&4.14e+02&3.55e+00\\
	\cmidrule(){2-12}
	&\multirow{3}*{BCKT}&mean&\cellcolor[rgb]{0.74,0.74,0.74}7.65e+00&1.99e+03&4.04e+07&7.49e+00&3.36e+01&\cellcolor[rgb]{0.74,0.74,0.74}1.73e+01&5.25e+02&5.87e+02&1.39e+01\\
	\cmidrule(){3-12}	&&std&1.42e+00&6.86e+02&2.11e+07&2.63e+00&1.34e+01&1.90e+00&2.61e+02&4.99e+02&3.88e+00\\
	\midrule
	\multirow{6}*{GL-SADE}&\multirow{3}*{/}&mean&1.48e+01&1.76e+00&2.04e+07&9.85e-01&2.49e+02&1.85e+01&3.88e-01&3.92e+02&1.36e+01\\
	\cmidrule(){3-12}	&&std&2.00e+00&8.39e-01&1.02e+07&4.30e-02&4.08e+02&1.81e+00&2.32e-01&2.45e+02&3.56e+00\\
	\cmidrule(){2-12}
	&\multirow{3}*{BCKT}&mean&\cellcolor[rgb]{0.74,0.74,0.74}1.03e+01&\cellcolor[rgb]{0.74,0.74,0.74}1.37e+00&1.91e+07&9.76e-01&3.19e+02&\cellcolor[rgb]{0.74,0.74,0.74}1.64e+01&3.80e-01&4.18e+02&1.38e+01\\
	\cmidrule(){3-12}	&&std&4.87e+00&9.70e-01&1.49e+07&5.15e-02&4.60e+02&2.71e+00&2.18e-01&2.32e+02&2.19e+00\\
	\midrule
	\multirow{6}*{\makecell[l]{DDEA-\\MESS}}&\multirow{3}*{/}&mean&5.05e+00&1.37e+00&2.35e+07&9.62e-01&7.64e+01&1.25e+01&4.24e-01&1.24e+01&5.98e+00\\
	\cmidrule(){3-12}	&&std&1.14e+00&3.89e-01&1.21e+07&3.94e-02&1.26e+02&4.19e+00&1.86e-01&1.01e+01&3.04e+00\\
	\cmidrule(){2-12}
	&\multirow{3}*{BCKT}&mean&5.55e+00&1.47e+00&\cellcolor[rgb]{0.74,0.74,0.74}1.69e+07&9.59e-01&4.90e+01&\cellcolor[rgb]{0.74,0.74,0.74}9.04e+00&3.76e-01&1.49e+01&5.54e+00\\
	\cmidrule(){3-12}	&&std&2.70e+00&6.15e-01&9.20e+06&4.03e-02&7.24e+01&3.48e+00&1.54e-01&1.37e+01&1.75e+00\\
	\midrule
	\multirow{6}*{LSADE}&\multirow{3}*{/}&mean&3.65e+00&1.31e+00&3.34e+07&9.64e-01&3.92e+01&1.22e+01&3.96e-01&5.17e+01&4.21e+00\\
	\cmidrule(){3-12}	&&std&9.67e-01&4.44e-01&1.35e+07&5.54e-02&5.68e+01&3.50e+00&2.19e-01&4.45e+01&1.53e+00\\
	\cmidrule(){2-12}
	&\multirow{3}*{BCKT}&mean&\cellcolor[rgb]{0.74,0.74,0.74}3.12e+00&1.36e+00&\cellcolor[rgb]{0.74,0.74,0.74}1.85e+07&9.44e-01&2.44e+01&\cellcolor[rgb]{0.74,0.74,0.74}1.00e+01&3.43e-01&8.06e+01&3.99e+00\\
	\cmidrule(){3-12}	&&std&6.81e-01&4.39e-01&7.28e+06&7.33e-02&1.58e+01&2.45e+00&1.26e-01&1.13e+02&1.31e+00\\
	\midrule
	\multirow{6}*{AutoSAEA}&\multirow{3}*{/}&mean&2.51e+00&1.35e+00&6.83e+02&1.01e+00&9.27e+00&1.19e+01&3.43e-01&1.38e+01&2.58e+00\\
	\cmidrule(){3-12}	&&std&3.51e-01&6.31e-01&4.75e+02&1.85e-02&1.40e+01&3.47e+00&1.86e-01&1.01e+01&6.51e-01\\
	\cmidrule(){2-12}
	&\multirow{3}*{BCKT}&mean&\cellcolor[rgb]{0.74,0.74,0.74}1.63e+00&1.63e+00&8.35e+02&1.01e+00&\cellcolor[rgb]{0.74,0.74,0.74}4.54e+00&\cellcolor[rgb]{0.74,0.74,0.74}8.16e+00&4.08e-01&1.54e+01&2.66e+00\\
	\cmidrule(){3-12}	&&std&3.96e-01&7.49e-01&5.43e+02&3.67e-02&2.69e+00&2.71e+00&2.00e-01&8.50e+00&6.51e-01\\
	\bottomrule
	\end{tabular}
	\label{tab:multi-result1}
\end{table*}

\begin{table*}[ht]
	\caption{Optimization results of MSAS-BCKT against MSAS with the six backbone optimizers on $T_2$ of the 9 MTOPs. Mean and standard deviation (i.e., mean$\pm$std) are based on 30 independent runs. The winner in each comparison between MSAS and MSAS-BCKT is highlighted based on the Wilcoxon rank-sum test with Holm $p$-value correction ($\alpha=0.05$).}
	\centering
	\heavyrulewidth=0.12em
	\lightrulewidth=0.08em
	\cmidrulewidth=0.05em
	%\fontsize{5.5pt}{5.5pt}\selectfont
	\scriptsize
	\setlength\tabcolsep{4.4pt}
\begin{tabular}{*{12}{lcccccccccccc}}
	\toprule
	\multirow{3}*{Optimizer}&\multirow{3}*{Transfer}&\multirow{3}*{Stat.}&\multicolumn{3}{c}{HS Problems}&\multicolumn{3}{c}{MS Problems}&\multicolumn{3}{c}{LS Problems}\\
	\cmidrule(l){4-6}\cmidrule(l){7-9}\cmidrule(l){10-12}
	&&&MTOP 1&MTOP 2&MTOP 3&MTOP 4&MTOP 5&MTOP 6&MTOP 7&MTOP 8&MTOP 9\\
	\midrule
	\multirow{6}*{BO-LCB}&\multirow{3}*{/}&mean&7.42e-01&1.21e+02&1.30e+07&1.61e+01&7.51e+00&9.45e+01&6.31e+01&6.95e+01&6.52e-01\\
	\cmidrule(){3-12}	&&std&2.02e-01&1.42e+01&1.18e+07&3.70e+00&6.79e+00&4.26e+01&1.41e+01&1.89e+02&1.56e-01\\
	\cmidrule(){2-12}
	&\multirow{3}*{BCKT}&mean&7.49e-01&\cellcolor[rgb]{0.74,0.74,0.74}9.01e+01&1.12e+07&\cellcolor[rgb]{0.74,0.74,0.74}1.03e+01&8.43e+00&9.80e+01&6.55e+01&3.77e+01&6.76e-01\\
	\cmidrule(){3-12}	&&std&2.33e-01&2.72e+01&8.35e+06&4.17e+00&4.63e+00&6.49e+01&1.63e+01&3.75e+01&1.47e-01\\
	\midrule
	\multirow{6}*{TLRBF}&\multirow{3}*{/}&mean&1.46e+00&2.19e+02&2.69e+07&2.47e+01&2.33e+02&4.49e+03&2.35e+02&1.28e+03&1.73e+00\\
	\cmidrule(){3-12}	&&std&2.12e-01&5.02e+01&2.09e+07&2.61e+00&1.32e+02&1.51e+03&4.98e+01&4.08e+02&7.76e-01\\
	\cmidrule(){2-12}
	&\multirow{3}*{BCKT}&mean&\cellcolor[rgb]{0.74,0.74,0.74}1.36e+00&\cellcolor[rgb]{0.74,0.74,0.74}1.77e+02&2.23e+07&\cellcolor[rgb]{0.74,0.74,0.74}2.27e+01&2.18e+02&5.07e+03&2.33e+02&1.20e+03&1.66e+00\\
	\cmidrule(){3-12}	&&std&1.71e-01&3.73e+01&1.68e+07&2.51e+00&9.27e+01&1.93e+03&3.67e+01&3.62e+02&6.27e-01\\
	\midrule
	\multirow{6}*{GL-SADE}&\multirow{3}*{/}&mean&9.82e-01&1.45e+02&1.85e+07&2.47e+01&1.13e+02&3.93e+00&1.30e+02&4.46e+02&9.70e-01\\
	\cmidrule(){3-12}	&&std&7.31e-02&1.80e+01&9.02e+06&2.69e+00&7.63e+01&1.80e+00&2.48e+01&2.97e+02&1.80e-01\\
	\cmidrule(){2-12}
	&\multirow{3}*{BCKT}&mean&1.01e+00&1.22e+02&2.18e+07&\cellcolor[rgb]{0.74,0.74,0.74}2.07e+01&1.37e+02&3.66e+00&1.30e+02&5.15e+02&\cellcolor[rgb]{0.74,0.74,0.74}8.77e-01\\
	\cmidrule(){3-12}	&&std&5.38e-02&5.55e+01&2.14e+07&3.55e+00&8.52e+01&1.54e+00&3.46e+01&2.59e+02&1.66e-01\\
	\midrule
	\multirow{6}*{\makecell[l]{DDEA-\\MESS}}&\multirow{3}*{/}&mean&9.49e-01&1.46e+02&4.25e+05&2.41e+01&9.29e+01&1.86e+00&1.38e+02&6.65e+02&\cellcolor[rgb]{0.74,0.74,0.74}8.32e-01\\
	\cmidrule(){3-12}	&&std&7.54e-02&1.82e+01&3.35e+05&3.14e+00&7.55e+01&6.11e-01&3.99e+01&3.99e+02&1.65e-01\\
	\cmidrule(){2-12}
	&\multirow{3}*{BCKT}&mean&9.61e-01&\cellcolor[rgb]{0.74,0.74,0.74}1.13e+02&4.48e+05&\cellcolor[rgb]{0.74,0.74,0.74}1.77e+01&9.19e+01&2.04e+00&1.52e+02&6.13e+02&9.41e-01\\
	\cmidrule(){3-12}	&&std&7.17e-02&5.25e+01&3.29e+05&4.59e+00&5.80e+01&5.29e-01&3.15e+01&2.83e+02&1.17e-01\\
	\midrule
	\multirow{6}*{LSADE}&\multirow{3}*{/}&mean&9.03e-01&1.27e+02&2.23e+06&2.39e+01&1.47e+02&2.13e+00&1.23e+02&7.33e+02&8.67e-01\\
	\cmidrule(){3-12}	&&std&1.09e-01&1.80e+01&2.09e+06&2.94e+00&1.11e+02&5.68e-01&2.94e+01&4.59e+02&1.58e-01\\
	\cmidrule(){2-12}
	&\multirow{3}*{BCKT}&mean&9.41e-01&\cellcolor[rgb]{0.74,0.74,0.74}8.71e+00&1.75e+06&\cellcolor[rgb]{0.74,0.74,0.74}1.69e+01&\cellcolor[rgb]{0.74,0.74,0.74}8.99e+01&2.24e+00&1.17e+02&6.52e+02&8.51e-01\\
	\cmidrule(){3-12}	&&std&7.11e-02&2.29e+01&1.05e+06&1.92e+00&8.27e+01&6.23e-01&2.22e+01&3.54e+02&1.81e-01\\
	\midrule
	\multirow{6}*{AutoSAEA}&\multirow{3}*{/}&mean&8.25e-01&1.15e+02&1.49e+06&1.99e+01&3.91e+00&2.33e+00&7.58e+01&3.68e+02&6.67e-01\\
	\cmidrule(){3-12}	&&std&1.13e-01&1.40e+01&1.03e+06&3.05e+00&6.94e+00&9.14e-01&2.30e+01&3.20e+02&1.67e-01\\
	\cmidrule(){2-12}
	&\multirow{3}*{BCKT}&mean&8.25e-01&\cellcolor[rgb]{0.74,0.74,0.74}1.38e+01&1.01e+06&\cellcolor[rgb]{0.74,0.74,0.74}1.45e+01&4.13e+00&2.07e+00&7.72e+01&2.68e+02&6.50e-01\\
	\cmidrule(){3-12}	&&std&9.10e-02&2.84e+01&6.12e+05&3.30e+00&7.71e+00&9.49e-01&1.70e+01&2.52e+02&1.94e-01\\
	\bottomrule
	\end{tabular}
	\label{tab:multi-result2}
\end{table*}

\begin{table*}[ht]
	\caption{Optimization results of MSAS-BCKT against MSAS with the six backbone optimizers on $T_1$ of the 6 MaTOPs. Mean and standard deviation (i.e., mean$\pm$std) are based on 30 independent runs. The winner in each comparison between MSAS and MSAS-BCKT is highlighted based on the Wilcoxon rank-sum test with Holm $p$-value correction ($\alpha=0.05$).}
	\centering
	\renewcommand{\arraystretch}{0.5}
	\heavyrulewidth=0.12em
	\lightrulewidth=0.08em
	\cmidrulewidth=0.05em
	%\fontsize{5.5pt}{5.5pt}\selectfont
	\scriptsize
	\setlength\tabcolsep{4.4pt}
\begin{tabular}{*{9}{lcccccccccccc}}
	\toprule
	\multirow{3}*{Optimizer}&\multirow{3}*{Transfer}&\multirow{3}*{Stat.}&\multicolumn{2}{c}{HS Problems}&\multicolumn{2}{c}{MS Problems}&\multicolumn{2}{c}{LS Problems}\\
	\cmidrule(l){4-5}\cmidrule(l){6-7}\cmidrule(l){8-9}
	&&&MaTOP 1&MaTOP 2&MaTOP 3&MaTOP 4&MaTOP 5&MaTOP 6\\
	\midrule
	\multirow{9}*{BO-LCB}&\multirow{3}*{/}&mean&2.80e+01&1.99e+01&1.08e+01&8.24e+01&1.56e+08&4.68e+00\\
	\cmidrule(){3-9}	&&std&8.93e+00&1.19e+00&5.06e+00&5.99e+01&3.68e+08&1.84e+00\\
	\cmidrule(){2-9}
	&\multirow{3}*{BCKT}&mean&2.89e+01&\cellcolor[rgb]{0.74,0.74,0.74}3.32e+00&9.02e+00&6.20e+01&5.76e+07&6.58e+00\\
	\cmidrule(){3-9}	&&std&9.90e+00&1.75e-01&4.41e+00&4.59e+01&9.56e+07&5.07e+00\\
	\midrule
	\multirow{9}*{TLRBF}&\multirow{3}*{/}&mean&2.11e+03&2.06e+01&3.49e+02&6.41e+02&5.26e+05&3.68e+02\\
	\cmidrule(){3-9}	&&std&1.36e+03&4.17e-01&1.91e+02&2.68e+02&1.36e+06&1.43e+02\\
	\cmidrule(){2-9}
	&\multirow{3}*{BCKT}&mean&1.89e+03&\cellcolor[rgb]{0.74,0.74,0.74}1.55e+01&4.92e+02&6.73e+02&1.45e+05&4.26e+02\\
	\cmidrule(){3-9}	&&std&8.67e+02&2.45e+00&3.63e+02&2.45e+02&6.31e+05&2.42e+02\\
	\midrule
	\multirow{9}*{GL-SADE}&\multirow{3}*{/}&mean&1.94e+00&1.90e+01&9.75e-01&2.51e+02&8.40e+07&1.08e+00\\
	\cmidrule(){3-9}	&&std&8.68e-01&2.23e+00&7.34e-01&1.49e+02&2.07e+08&7.06e-01\\
	\cmidrule(){2-9}
	&\multirow{3}*{BCKT}&mean&1.43e+00&\cellcolor[rgb]{0.74,0.74,0.74}1.24e+01&9.62e-01&2.48e+02&1.07e+07&7.33e-01\\
	\cmidrule(){3-9}	&&std&6.37e-01&2.02e+00&4.49e-01&1.31e+02&3.27e+07&5.68e-01\\
	\midrule
	\multirow{9}*{\makecell[l]{DDEA-\\MESS}}&\multirow{3}*{/}&mean&1.39e+00&1.85e+01&1.08e+00&2.69e+02&1.28e+05&1.04e+00\\
	\cmidrule(){3-9}	&&std&4.93e-01&3.00e+00&5.05e-01&2.07e+02&4.17e+05&3.97e-01\\
	\cmidrule(){2-9}
	&\multirow{3}*{BCKT}&mean&1.24e+00&\cellcolor[rgb]{0.74,0.74,0.74}6.11e+00&1.06e+00&3.25e+02&6.30e+05&9.91e-01\\
	\cmidrule(){3-9}	&&std&5.23e-01&3.54e+00&3.93e-01&2.61e+02&2.70e+06&3.58e-01\\
	\midrule
	\multirow{9}*{LSADE}&\multirow{3}*{/}&mean&1.54e+00&1.59e+01&9.54e-01&3.97e+02&9.18e+06&1.02e+00\\
	\cmidrule(){3-9}	&&std&6.80e-01&4.13e+00&3.69e-01&3.03e+02&2.85e+07&4.16e-01\\
	\cmidrule(){2-9}
	&\multirow{3}*{BCKT}&mean&1.45e+00&\cellcolor[rgb]{0.74,0.74,0.74}1.93e+00&9.78e-01&4.36e+02&\cellcolor[rgb]{0.74,0.74,0.74}5.62e+04&1.05e+00\\
	\cmidrule(){3-9}	&&std&4.12e-01&8.84e-01&4.66e-01&3.28e+02&1.41e+05&5.35e-01\\
	\midrule
	\multirow{9}*{AutoSAEA}&\multirow{3}*{/}&mean&8.90e-01&1.30e+01&9.84e-01&5.87e+01&6.08e+02&9.72e-01\\
	\cmidrule(){3-9}	&&std&6.52e-01&4.10e+00&3.73e-01&7.58e+01&2.42e+03&6.86e-01\\
	\cmidrule(){2-9}
	&\multirow{3}*{BCKT}&mean&1.17e+00&\cellcolor[rgb]{0.74,0.74,0.74}3.36e+00&1.02e+00&4.83e+01&1.12e+02&1.06e+00\\
	\cmidrule(){3-9}	&&std&7.57e-01&3.88e+00&5.30e-01&5.77e+01&1.66e+02&5.94e-01\\
	\bottomrule
	\end{tabular}
	\label{tab:many-result1}
\end{table*}

\begin{table*}[ht]
	\caption{Optimization results of MSAS-BCKT against MSAS with the six backbone optimizers on $T_2$ of the 6 MaTOPs. Mean and standard deviation (i.e., mean$\pm$std) are based on 30 independent runs. The winner in each comparison between MSAS and MSAS-BCKT is highlighted based on the Wilcoxon rank-sum test with Holm $p$-value correction ($\alpha=0.05$).}
	\centering
	\renewcommand{\arraystretch}{0.5}
	\heavyrulewidth=0.12em
	\lightrulewidth=0.08em
	\cmidrulewidth=0.05em
	%\fontsize{5.5pt}{5.5pt}\selectfont
	\scriptsize
	\setlength\tabcolsep{4.4pt}
\begin{tabular}{*{9}{lcccccccccccc}}
	\toprule
	\multirow{3}*{Optimizer}&\multirow{3}*{Transfer}&\multirow{3}*{Stat.}&\multicolumn{2}{c}{HS Problems}&\multicolumn{2}{c}{MS Problems}&\multicolumn{2}{c}{LS Problems}\\
	\cmidrule(l){4-5}\cmidrule(l){6-7}\cmidrule(l){8-9}
	&&&MaTOP 1&MaTOP 2&MaTOP 3&MaTOP 4&MaTOP 5&MaTOP 6\\
	\midrule
	\multirow{9}*{BO-LCB}&\multirow{3}*{/}&mean&1.26e+02&1.00e+02&4.02e+01&1.50e+01&7.40e+01&3.34e+01\\
	\cmidrule(){3-9}	&&std&1.25e+01&7.03e+01&7.87e+00&7.06e+00&4.56e+01&6.53e+00\\
	\cmidrule(){2-9}
	&\multirow{3}*{BCKT}&mean&\cellcolor[rgb]{0.74,0.74,0.74}9.79e+01&8.88e+01&3.48e+01&\cellcolor[rgb]{0.74,0.74,0.74}3.68e+00&8.96e+01&3.41e+01\\
	\cmidrule(){3-9}	&&std&1.81e+01&6.23e+01&7.40e+00&5.92e-01&9.22e+01&6.47e+00\\
	\midrule
	\multirow{9}*{TLRBF}&\multirow{3}*{/}&mean&2.10e+02&6.30e+03&1.23e+02&1.95e+01&8.62e+02&5.24e+01\\
	\cmidrule(){3-9}	&&std&4.71e+01&1.77e+03&3.44e+01&2.14e+00&2.76e+02&8.19e+01\\
	\cmidrule(){2-9}
	&\multirow{3}*{BCKT}&mean&\cellcolor[rgb]{0.74,0.74,0.74}1.56e+02&5.88e+03&1.06e+02&\cellcolor[rgb]{0.74,0.74,0.74}1.48e+01&8.53e+02&3.66e+01\\
	\cmidrule(){3-9}	&&std&2.74e+01&2.26e+03&2.28e+01&2.94e+00&3.01e+02&1.33e+01\\
	\midrule
	\multirow{9}*{GL-SADE}&\multirow{3}*{/}&mean&1.46e+02&3.88e+00&7.59e+01&1.83e+01&3.30e+02&3.52e+01\\
	\cmidrule(){3-9}	&&std&2.28e+01&1.53e+00&2.23e+01&2.58e+00&1.46e+02&9.11e+00\\
	\cmidrule(){2-9}
	&\multirow{3}*{BCKT}&mean&1.27e+02&4.78e+00&7.51e+01&\cellcolor[rgb]{0.74,0.74,0.74}1.60e+01&4.29e+02&3.50e+01\\
	\cmidrule(){3-9}	&&std&5.67e+01&1.64e+00&1.59e+01&1.98e+00&2.69e+02&6.69e+00\\
	\midrule
	\multirow{9}*{\makecell[l]{DDEA-\\MESS}}&\multirow{3}*{/}&mean&1.41e+02&2.15e+00&8.14e+01&1.11e+01&4.64e+02&9.87e+00\\
	\cmidrule(){3-9}	&&std&2.09e+01&8.53e-01&1.70e+01&5.99e+00&2.20e+02&5.86e+00\\
	\cmidrule(){2-9}
	&\multirow{3}*{BCKT}&mean&1.21e+02&1.95e+00&7.30e+01&8.20e+00&4.54e+02&1.10e+01\\
	\cmidrule(){3-9}	&&std&4.81e+01&5.98e-01&1.64e+01&1.62e+00&3.11e+02&7.42e+00\\
	\midrule
	\multirow{9}*{LSADE}&\multirow{3}*{/}&mean&1.33e+02&2.01e+00&7.43e+01&1.15e+01&3.32e+02&5.97e+00\\
	\cmidrule(){3-9}	&&std&2.56e+01&5.62e-01&2.27e+01&6.28e+00&1.95e+02&3.91e+00\\
	\cmidrule(){2-9}
	&\multirow{3}*{BCKT}&mean&\cellcolor[rgb]{0.74,0.74,0.74}3.43e+01&2.34e+00&6.49e+01&8.22e+00&3.31e+02&6.11e+00\\
	\cmidrule(){3-9}	&&std&5.15e+01&9.52e-01&1.60e+01&2.54e+00&2.10e+02&5.02e+00\\
	\midrule
	\multirow{9}*{AutoSAEA}&\multirow{3}*{/}&mean&1.18e+02&1.92e+00&4.00e+01&8.56e+00&3.69e+02&\cellcolor[rgb]{0.74,0.74,0.74}9.62e-01\\
	\cmidrule(){3-9}	&&std&1.17e+01&1.01e+00&1.56e+01&5.19e+00&3.11e+02&2.65e-01\\
	\cmidrule(){2-9}
	&\multirow{3}*{BCKT}&mean&\cellcolor[rgb]{0.74,0.74,0.74}4.47e+01&2.08e+00&3.81e+01&\cellcolor[rgb]{0.74,0.74,0.74}5.03e+00&2.82e+02&1.53e+00\\
	\cmidrule(){3-9}	&&std&4.75e+01&8.95e-01&1.31e+01&7.21e-01&2.78e+02&6.55e-01\\
	\bottomrule
	\end{tabular}
	\label{tab:many-result2}
\end{table*}

\begin{table*}[ht]
	\caption{Optimization results of MSAS-BCKT against MSAS with the six backbone optimizers on $T_3$ of the 6 MaTOPs. Mean and standard deviation (i.e., mean$\pm$std) are based on 30 independent runs. The winner in each comparison between MSAS and MSAS-BCKT is highlighted based on the Wilcoxon rank-sum test with Holm $p$-value correction ($\alpha=0.05$).}
	\centering
	\renewcommand{\arraystretch}{0.5}
	\heavyrulewidth=0.12em
	\lightrulewidth=0.08em
	\cmidrulewidth=0.05em
	%\fontsize{5.5pt}{5.5pt}\selectfont
	\scriptsize
	\setlength\tabcolsep{4.4pt}
\begin{tabular}{*{9}{lcccccccccccc}}
	\toprule
	\multirow{3}*{Optimizer}&\multirow{3}*{Transfer}&\multirow{3}*{Stat.}&\multicolumn{2}{c}{HS Problems}&\multicolumn{2}{c}{MS Problems}&\multicolumn{2}{c}{LS Problems}\\
	\cmidrule(l){4-5}\cmidrule(l){6-7}\cmidrule(l){8-9}
	&&&MaTOP 1&MaTOP 2&MaTOP 3&MaTOP 4&MaTOP 5&MaTOP 6\\
	\midrule
	\multirow{9}*{BO-LCB}&\multirow{3}*{/}&mean&1.31e+01&1.59e+07&1.87e+04&\cellcolor[rgb]{0.74,0.74,0.74}2.86e+01&3.59e+02&5.86e+00\\
	\cmidrule(){3-9}	&&std&6.69e+00&9.97e+06&9.53e+02&2.67e+01&2.64e+02&2.34e+00\\
	\cmidrule(){2-9}
	&\multirow{3}*{BCKT}&mean&\cellcolor[rgb]{0.74,0.74,0.74}3.24e+00&1.63e+07&1.86e+04&4.34e+01&4.93e+02&6.99e+00\\
	\cmidrule(){3-9}	&&std&2.45e-01&1.06e+07&1.08e+03&2.94e+01&5.23e+02&1.94e+00\\
	\midrule
	\multirow{9}*{TLRBF}&\multirow{3}*{/}&mean&1.87e+01&1.35e+07&5.60e+06&1.74e+03&8.84e+02&9.10e+00\\
	\cmidrule(){3-9}	&&std&1.97e+00&7.91e+06&2.59e+06&6.25e+02&7.29e+02&1.68e+00\\
	\cmidrule(){2-9}
	&\multirow{3}*{BCKT}&mean&\cellcolor[rgb]{0.74,0.74,0.74}1.18e+01&2.40e+07&5.85e+06&1.73e+03&1.15e+03&9.61e+00\\
	\cmidrule(){3-9}	&&std&2.23e+00&1.82e+07&3.01e+06&8.01e+02&8.53e+02&1.76e+00\\
	\midrule
	\multirow{9}*{GL-SADE}&\multirow{3}*{/}&mean&1.56e+01&1.49e+07&1.90e+06&1.61e+00&7.18e+02&1.01e+01\\
	\cmidrule(){3-9}	&&std&3.10e+00&7.74e+06&1.08e+06&1.00e+00&3.33e+02&1.11e+00\\
	\cmidrule(){2-9}
	&\multirow{3}*{BCKT}&mean&\cellcolor[rgb]{0.74,0.74,0.74}9.20e+00&1.71e+07&1.80e+06&1.87e+00&8.55e+02&1.05e+01\\
	\cmidrule(){3-9}	&&std&3.79e+00&6.18e+06&1.08e+06&9.39e-01&5.40e+02&1.88e+00\\
	\midrule
	\multirow{9}*{\makecell[l]{DDEA-\\MESS}}&\multirow{3}*{/}&mean&8.34e+00&5.54e+05&3.65e+06&1.43e+00&\cellcolor[rgb]{0.74,0.74,0.74}2.42e+01&8.06e+00\\
	\cmidrule(){3-9}	&&std&3.42e+00&5.58e+05&1.98e+06&3.88e-01&4.05e+01&1.84e+00\\
	\cmidrule(){2-9}
	&\multirow{3}*{BCKT}&mean&\cellcolor[rgb]{0.74,0.74,0.74}6.32e+00&3.96e+05&2.94e+06&1.38e+00&3.07e+01&8.28e+00\\
	\cmidrule(){3-9}	&&std&3.23e+00&3.40e+05&1.66e+06&6.33e-01&2.61e+01&2.24e+00\\
	\midrule
	\multirow{9}*{LSADE}&\multirow{3}*{/}&mean&9.99e+00&2.52e+06&3.78e+06&1.63e+00&9.25e+01&8.60e+00\\
	\cmidrule(){3-9}	&&std&5.74e+00&2.66e+06&2.15e+06&5.76e-01&4.99e+01&1.97e+00\\
	\cmidrule(){2-9}
	&\multirow{3}*{BCKT}&mean&\cellcolor[rgb]{0.74,0.74,0.74}2.57e+00&2.72e+06&3.69e+06&1.68e+00&9.34e+01&8.58e+00\\
	\cmidrule(){3-9}	&&std&4.43e-01&2.04e+06&1.74e+06&4.99e-01&6.48e+01&2.50e+00\\
	\midrule
	\multirow{9}*{AutoSAEA}&\multirow{3}*{/}&mean&6.53e+00&1.03e+06&2.45e+00&1.07e+00&1.54e+01&3.77e+00\\
	\cmidrule(){3-9}	&&std&2.40e+00&5.95e+05&2.91e+00&5.86e-01&1.17e+01&1.04e+00\\
	\cmidrule(){2-9}
	&\multirow{3}*{BCKT}&mean&\cellcolor[rgb]{0.74,0.74,0.74}2.67e+00&1.55e+06&2.11e+00&1.27e+00&1.91e+01&4.22e+00\\
	\cmidrule(){3-9}	&&std&1.15e+00&1.15e+06&1.02e+00&9.60e-01&1.31e+01&2.00e+00\\
	\bottomrule
	\end{tabular}
	\label{tab:many-result3}
\end{table*}

\begin{table*}[ht]
	\caption{Optimization results of MSAS-BCKT against MSAS with the six backbone optimizers on $T_4$ of the 6 MaTOPs. Mean and standard deviation (i.e., mean$\pm$std) are based on 30 independent runs. The winner in each comparison between MSAS and MSAS-BCKT is highlighted based on the Wilcoxon rank-sum test with Holm $p$-value correction ($\alpha=0.05$).}
	\centering
	\renewcommand{\arraystretch}{0.5}
	\heavyrulewidth=0.12em
	\lightrulewidth=0.08em
	\cmidrulewidth=0.05em
	%\fontsize{5.5pt}{5.5pt}\selectfont
	\scriptsize
	\setlength\tabcolsep{4.4pt}
\begin{tabular}{*{9}{lcccccccccccc}}
	\toprule
	\multirow{3}*{Optimizer}&\multirow{3}*{Transfer}&\multirow{3}*{Stat.}&\multicolumn{2}{c}{HS Problems}&\multicolumn{2}{c}{MS Problems}&\multicolumn{2}{c}{LS Problems}\\
	\cmidrule(l){4-5}\cmidrule(l){6-7}\cmidrule(l){8-9}
	&&&MaTOP 1&MaTOP 2&MaTOP 3&MaTOP 4&MaTOP 5&MaTOP 6\\
	\midrule
	\multirow{9}*{BO-LCB}&\multirow{3}*{/}&mean&1.77e+06&1.85e+01&5.09e+00&1.25e+02&1.25e+02&8.01e-01\\
	\cmidrule(){3-9}	&&std&3.89e+05&3.12e+00&2.15e+00&8.22e+00&1.12e+01&1.49e-01\\
	\cmidrule(){2-9}
	&\multirow{3}*{BCKT}&mean&\cellcolor[rgb]{0.74,0.74,0.74}9.73e+05&\cellcolor[rgb]{0.74,0.74,0.74}4.87e+00&3.66e+00&1.22e+02&1.18e+02&8.60e-01\\
	\cmidrule(){3-9}	&&std&5.99e+05&5.41e-01&2.04e+00&1.26e+01&1.80e+01&1.22e-01\\
	\midrule
	\multirow{9}*{TLRBF}&\multirow{3}*{/}&mean&2.35e+07&2.50e+01&8.09e+00&2.08e+02&3.62e+02&1.58e+00\\
	\cmidrule(){3-9}	&&std&1.29e+07&2.25e+00&1.86e+00&2.86e+01&6.13e+01&3.16e-01\\
	\cmidrule(){2-9}
	&\multirow{3}*{BCKT}&mean&\cellcolor[rgb]{0.74,0.74,0.74}1.22e+07&\cellcolor[rgb]{0.74,0.74,0.74}2.10e+01&7.90e+00&1.96e+02&3.38e+02&1.62e+00\\
	\cmidrule(){3-9}	&&std&5.26e+06&2.95e+00&1.59e+00&2.19e+01&6.86e+01&3.14e-01\\
	\midrule
	\multirow{9}*{GL-SADE}&\multirow{3}*{/}&mean&8.31e+06&2.52e+01&7.68e+00&1.50e+02&1.82e+02&1.01e+00\\
	\cmidrule(){3-9}	&&std&4.94e+06&2.72e+00&1.44e+00&2.11e+01&3.62e+01&1.15e-01\\
	\cmidrule(){2-9}
	&\multirow{3}*{BCKT}&mean&\cellcolor[rgb]{0.74,0.74,0.74}1.12e+06&\cellcolor[rgb]{0.74,0.74,0.74}1.59e+01&7.66e+00&1.56e+02&1.92e+02&9.93e-01\\
	\cmidrule(){3-9}	&&std&1.12e+06&8.32e+00&1.37e+00&1.70e+01&3.21e+01&8.06e-02\\
	\midrule
	\multirow{9}*{\makecell[l]{DDEA-\\MESS}}&\multirow{3}*{/}&mean&1.04e+07&2.45e+01&7.10e+00&1.47e+02&1.96e+02&9.15e-01\\
	\cmidrule(){3-9}	&&std&4.28e+06&2.55e+00&1.61e+00&2.49e+01&3.62e+01&9.27e-02\\
	\cmidrule(){2-9}
	&\multirow{3}*{BCKT}&mean&\cellcolor[rgb]{0.74,0.74,0.74}2.27e+06&\cellcolor[rgb]{0.74,0.74,0.74}8.83e+00&6.69e+00&1.41e+02&2.12e+02&9.19e-01\\
	\cmidrule(){3-9}	&&std&3.84e+06&7.54e+00&1.78e+00&2.52e+01&4.73e+01&8.65e-02\\
	\midrule
	\multirow{9}*{LSADE}&\multirow{3}*{/}&mean&1.49e+07&2.52e+01&7.09e+00&1.31e+02&1.83e+02&8.77e-01\\
	\cmidrule(){3-9}	&&std&6.10e+06&2.36e+00&1.22e+00&2.16e+01&3.64e+01&1.24e-01\\
	\cmidrule(){2-9}
	&\multirow{3}*{BCKT}&mean&\cellcolor[rgb]{0.74,0.74,0.74}3.01e+05&\cellcolor[rgb]{0.74,0.74,0.74}5.95e+00&6.82e+00&1.30e+02&1.93e+02&9.57e-01\\
	\cmidrule(){3-9}	&&std&4.16e+05&7.51e+00&1.34e+00&1.48e+01&3.57e+01&1.76e-01\\
	\midrule
	\multirow{9}*{AutoSAEA}&\multirow{3}*{/}&mean&2.14e+01&2.03e+01&2.41e+00&1.19e+02&1.57e+02&8.26e-01\\
	\cmidrule(){3-9}	&&std&1.46e+01&3.75e+00&9.31e-01&1.32e+01&2.94e+01&1.20e-01\\
	\cmidrule(){2-9}
	&\multirow{3}*{BCKT}&mean&2.20e+01&\cellcolor[rgb]{0.74,0.74,0.74}6.52e+00&2.25e+00&1.15e+02&1.43e+02&7.51e-01\\
	\cmidrule(){3-9}	&&std&1.45e+01&4.01e+00&8.95e-01&1.33e+01&2.07e+01&1.39e-01\\
	\bottomrule
	\end{tabular}
	\label{tab:many-result4}
\end{table*}

\begin{table*}[ht]
	\caption{Optimization results of MSAS-BCKT against MSAS with the six backbone optimizers on $T_5$ of the 6 MaTOPs. Mean and standard deviation (i.e., mean$\pm$std) are based on 30 independent runs. The winner in each comparison between MSAS and MSAS-BCKT is highlighted based on the Wilcoxon rank-sum test with Holm $p$-value correction ($\alpha=0.05$).}
	\centering
	\renewcommand{\arraystretch}{0.5}
	\heavyrulewidth=0.12em
	\lightrulewidth=0.08em
	\cmidrulewidth=0.05em
	%\fontsize{5.5pt}{5.5pt}\selectfont
	\scriptsize
	\setlength\tabcolsep{4.4pt}
\begin{tabular}{*{9}{lcccccccccccc}}
	\toprule
	\multirow{3}*{Optimizer}&\multirow{3}*{Transfer}&\multirow{3}*{Stat.}&\multicolumn{2}{c}{HS Problems}&\multicolumn{2}{c}{MS Problems}&\multicolumn{2}{c}{LS Problems}\\
	\cmidrule(l){4-5}\cmidrule(l){6-7}\cmidrule(l){8-9}
	&&&MaTOP 1&MaTOP 2&MaTOP 3&MaTOP 4&MaTOP 5&MaTOP 6\\
	\midrule
	\multirow{9}*{BO-LCB}&\multirow{3}*{/}&mean&1.02e+00&1.09e+00&1.00e+02&9.72e-01&2.07e+01&8.48e+04\\
	\cmidrule(){3-9}	&&std&4.17e-02&4.12e-02&1.27e+02&8.44e-02&9.12e-02&8.55e+03\\
	\cmidrule(){2-9}
	&\multirow{3}*{BCKT}&mean&1.02e+00&1.09e+00&7.89e+01&9.96e-01&2.06e+01&8.57e+04\\
	\cmidrule(){3-9}	&&std&1.73e-02&5.21e-02&7.10e+01&6.20e-02&1.48e-01&8.38e+03\\
	\midrule
	\multirow{9}*{TLRBF}&\multirow{3}*{/}&mean&3.16e+00&6.94e+00&4.13e+01&3.40e+00&2.09e+01&5.00e+06\\
	\cmidrule(){3-9}	&&std&9.13e-01&2.08e+00&5.10e+01&7.06e-01&3.19e-01&2.48e+06\\
	\cmidrule(){2-9}
	&\multirow{3}*{BCKT}&mean&2.99e+00&7.07e+00&3.71e+01&3.62e+00&2.06e+01&5.12e+06\\
	\cmidrule(){3-9}	&&std&1.00e+00&2.22e+00&2.44e+01&8.62e-01&1.15e+00&2.81e+06\\
	\midrule
	\multirow{9}*{GL-SADE}&\multirow{3}*{/}&mean&9.83e-01&9.88e-01&1.34e+02&1.00e+00&2.01e+01&1.96e+06\\
	\cmidrule(){3-9}	&&std&6.97e-02&3.27e-02&1.38e+02&3.12e-02&1.51e+00&1.52e+06\\
	\cmidrule(){2-9}
	&\multirow{3}*{BCKT}&mean&1.00e+00&9.60e-01&1.55e+02&\cellcolor[rgb]{0.74,0.74,0.74}9.79e-01&1.94e+01&1.29e+06\\
	\cmidrule(){3-9}	&&std&1.36e-02&9.34e-02&1.31e+02&5.27e-02&2.61e+00&6.88e+05\\
	\midrule
	\multirow{9}*{\makecell[l]{DDEA-\\MESS}}&\multirow{3}*{/}&mean&9.57e-01&9.58e-01&4.66e+01&9.61e-01&2.01e+01&2.59e+06\\
	\cmidrule(){3-9}	&&std&4.27e-02&3.53e-02&4.86e+01&8.30e-02&2.03e+00&1.89e+06\\
	\cmidrule(){2-9}
	&\multirow{3}*{BCKT}&mean&9.54e-01&9.21e-01&2.99e+01&9.70e-01&1.99e+01&2.49e+06\\
	\cmidrule(){3-9}	&&std&5.85e-02&8.30e-02&1.13e+01&3.63e-02&2.61e+00&1.65e+06\\
	\midrule
	\multirow{9}*{LSADE}&\multirow{3}*{/}&mean&9.43e-01&9.57e-01&2.05e+01&9.68e-01&2.09e+01&2.10e+06\\
	\cmidrule(){3-9}	&&std&1.29e-01&5.77e-02&1.18e+01&7.09e-02&2.44e-01&1.21e+06\\
	\cmidrule(){2-9}
	&\multirow{3}*{BCKT}&mean&9.55e-01&9.63e-01&2.32e+01&9.42e-01&\cellcolor[rgb]{0.74,0.74,0.74}1.96e+01&2.39e+06\\
	\cmidrule(){3-9}	&&std&5.03e-02&4.93e-02&1.06e+01&9.34e-02&2.16e+00&1.32e+06\\
	\midrule
	\multirow{9}*{AutoSAEA}&\multirow{3}*{/}&mean&9.37e-01&1.02e+00&3.68e+00&9.49e-01&1.83e+01&8.11e+00\\
	\cmidrule(){3-9}	&&std&7.79e-02&1.21e-02&2.41e+00&4.78e-02&4.20e+00&8.47e+00\\
	\cmidrule(){2-9}
	&\multirow{3}*{BCKT}&mean&9.47e-01&1.02e+00&5.49e+00&9.37e-01&1.95e+01&3.61e+00\\
	\cmidrule(){3-9}	&&std&5.87e-02&1.12e-02&4.20e+00&5.94e-02&2.99e+00&3.20e+00\\
	\bottomrule
	\end{tabular}
	\label{tab:many-result5}
\end{table*}

\begin{table*}[ht]
	\caption{Average objective values and their standard deviation obtained by MaTDE, MFEA-II, MMaTEA-DGT, LCB-EMT, RAMTEA, SELF and BO-LCB-BCKT on the 9 MTOPs over 30 runs. The highlighted entries are significantly better based on the Wilcoxon rank-sum test with Holm $p$-value correction ($\alpha=0.05$).}
	\centering
	\renewcommand{\arraystretch}{0.5}
	\heavyrulewidth=0.12em
	\lightrulewidth=0.08em
	\cmidrulewidth=0.05em
	%\fontsize{5.5pt}{5.5pt}\selectfont
	\scriptsize
	\setlength\tabcolsep{4.4pt}
\begin{tabular}{*{9}{lcccccccccccc}}
	\toprule
	{Problem}&{Task}&\multicolumn{1}{c}{MaTDE}&\multicolumn{1}{c}{MFEA-II}&\multicolumn{1}{c}{MMaTEA-DGT}&\multicolumn{1}{c}{LCB-EMT}&\multicolumn{1}{c}{RAMTEA}&\multicolumn{1}{c}{SELF}&\multicolumn{1}{c}{BO-LCB-BCKT}\\
	\midrule
	\multirow{4}*{MTOP 1}&{$T_1$}&1.46e+01$\pm$1.98e+00&1.88e+01$\pm$1.16e+00&1.87e+01$\pm$1.44e+00&1.54e+01$\pm$1.93e+00&1.96e+01$\pm$7.49e-01&3.12e+00$\pm$1.23e+00&\cellcolor[rgb]{0.74,0.74,0.74}1.97e+00$\pm$4.00e-01\\
	\cmidrule(){2-9}
	&{$T_2$}&3.73e+00$\pm$1.43e+00&8.24e+00$\pm$4.87e+00&1.11e+01$\pm$3.61e+00&5.19e+00$\pm$2.27e+00&1.51e+01$\pm$4.34e+00&\cellcolor[rgb]{0.74,0.74,0.74}6.90e-01$\pm$2.43e-01&\cellcolor[rgb]{0.74,0.74,0.74}7.49e-01$\pm$2.33e-01\\
	\midrule
	\multirow{4}*{MTOP 2}&{$T_1$}&1.34e+04$\pm$4.68e+03&2.60e+04$\pm$1.01e+04&2.76e+04$\pm$8.98e+03&1.50e+04$\pm$6.20e+03&3.82e+04$\pm$1.15e+04&6.62e+02$\pm$4.02e+02&\cellcolor[rgb]{0.74,0.74,0.74}3.38e+01$\pm$3.19e+01\\
	\cmidrule(){2-9}
	&{$T_2$}&2.58e+02$\pm$4.38e+01&3.91e+02$\pm$1.09e+02&4.82e+02$\pm$1.32e+02&2.73e+02$\pm$8.44e+01&5.27e+02$\pm$1.10e+02&\cellcolor[rgb]{0.74,0.74,0.74}5.59e+01$\pm$1.49e+01&9.01e+01$\pm$2.72e+01\\
	\midrule
	\multirow{4}*{MTOP 3}&{$T_1$}&\cellcolor[rgb]{0.74,0.74,0.74}6.04e+07$\pm$2.61e+07&2.77e+08$\pm$2.41e+08&1.89e+08$\pm$1.02e+08&2.24e+08$\pm$1.63e+08&3.57e+08$\pm$1.33e+08&\cellcolor[rgb]{0.74,0.74,0.74}5.57e+07$\pm$5.24e+07&9.44e+07$\pm$3.45e+07\\
	\cmidrule(){2-9}
	&{$T_2$}&4.08e+08$\pm$1.56e+08&1.27e+09$\pm$8.02e+08&1.53e+09$\pm$8.59e+08&1.40e+09$\pm$8.42e+08&2.45e+09$\pm$8.44e+08&1.10e+08$\pm$1.13e+08&\cellcolor[rgb]{0.74,0.74,0.74}1.12e+07$\pm$8.35e+06\\
	\midrule
	\multirow{4}*{\makecell[l]{MTOP 4}}&{$T_1$}&2.47e+01$\pm$6.69e+00&3.79e+01$\pm$1.04e+01&5.33e+01$\pm$1.40e+01&3.75e+01$\pm$1.07e+01&6.18e+01$\pm$1.47e+01&6.56e+00$\pm$3.78e+00&\cellcolor[rgb]{0.74,0.74,0.74}1.14e+00$\pm$2.86e-02\\
	\cmidrule(){2-9}
	&{$T_2$}&2.43e+01$\pm$1.67e+00&2.47e+01$\pm$2.23e+00&2.95e+01$\pm$1.44e+00&2.34e+01$\pm$2.67e+00&3.04e+01$\pm$1.70e+00&\cellcolor[rgb]{0.74,0.74,0.74}4.39e+00$\pm$2.18e+00&1.03e+01$\pm$4.17e+00\\
	\midrule
	\multirow{4}*{MTOP 5}&{$T_1$}&\cellcolor[rgb]{0.74,0.74,0.74}1.62e+01$\pm$6.61e+00&4.80e+01$\pm$1.07e+02&3.76e+02$\pm$7.55e+02&2.37e+01$\pm$8.13e+00&9.48e+01$\pm$1.09e+02&4.14e+03$\pm$5.39e+00&6.52e+01$\pm$4.18e+01\\
	\cmidrule(){2-9}
	&{$T_2$}&1.52e+02$\pm$8.03e+01&2.20e+02$\pm$8.62e+01&4.72e+02$\pm$2.32e+02&2.13e+02$\pm$9.57e+01&5.11e+02$\pm$1.70e+02&5.84e+01$\pm$4.20e+01&\cellcolor[rgb]{0.74,0.74,0.74}8.43e+00$\pm$4.63e+00\\
	\midrule
	\multirow{4}*{MTOP 6}&{$T_1$}&1.93e+01$\pm$6.79e-01&2.02e+01$\pm$4.59e-01&2.08e+01$\pm$3.62e-01&1.99e+01$\pm$6.80e-01&2.08e+01$\pm$2.02e-01&1.48e+01$\pm$1.60e+00&\cellcolor[rgb]{0.74,0.74,0.74}5.63e+00$\pm$1.83e+00\\
	\cmidrule(){2-9}
	&{$T_2$}&2.00e+04$\pm$5.77e+03&3.49e+04$\pm$9.68e+03&4.46e+04$\pm$1.26e+04&2.87e+04$\pm$9.15e+03&5.19e+04$\pm$9.28e+03&5.08e+03$\pm$2.76e+03&\cellcolor[rgb]{0.74,0.74,0.74}9.80e+01$\pm$6.49e+01\\
	\midrule
	\multirow{4}*{MTOP 7}&{$T_1$}&5.05e+03$\pm$1.31e+03&6.98e+03$\pm$2.18e+03&7.34e+03$\pm$2.40e+03&6.16e+03$\pm$2.31e+03&1.14e+04$\pm$3.42e+03&7.01e+02$\pm$6.43e+02&\cellcolor[rgb]{0.74,0.74,0.74}2.16e+01$\pm$6.31e+00\\
	\cmidrule(){2-9}
	&{$T_2$}&3.85e+02$\pm$6.63e+01&4.55e+02$\pm$1.06e+02&5.46e+02$\pm$1.47e+02&4.54e+02$\pm$1.07e+02&7.44e+02$\pm$1.24e+02&1.14e+02$\pm$4.49e+01&\cellcolor[rgb]{0.74,0.74,0.74}6.55e+01$\pm$1.63e+01\\
	\midrule
	\multirow{4}*{MTOP 8}&{$T_1$}&6.88e+03$\pm$4.04e+03&1.80e+04$\pm$1.24e+04&1.76e+04$\pm$9.79e+03&1.17e+04$\pm$5.31e+03&3.04e+04$\pm$1.11e+04&8.62e+07$\pm$6.55e+07&\cellcolor[rgb]{0.74,0.74,0.74}2.65e+02$\pm$1.39e+02\\
	\cmidrule(){2-9}
	&{$T_2$}&1.23e+03$\pm$3.53e+02&1.06e+03$\pm$3.29e+02&2.26e+03$\pm$5.68e+02&1.09e+03$\pm$3.89e+02&2.36e+03$\pm$4.75e+02&4.55e+03$\pm$1.25e+02&\cellcolor[rgb]{0.74,0.74,0.74}3.77e+01$\pm$3.75e+01\\
	\midrule
	\multirow{4}*{MTOP 9}&{$T_1$}&1.58e+01$\pm$1.70e+00&1.81e+01$\pm$1.39e+00&1.89e+01$\pm$1.64e+00&1.75e+01$\pm$1.47e+00&2.01e+01$\pm$5.38e-01&1.15e+01$\pm$4.43e+00&\cellcolor[rgb]{0.74,0.74,0.74}3.69e+00$\pm$3.33e+00\\
	\cmidrule(){2-9}
	&{$T_2$}&1.94e+01$\pm$5.01e+00&2.38e+01$\pm$1.02e+01&2.51e+01$\pm$9.19e+00&1.95e+01$\pm$9.87e+00&4.14e+01$\pm$7.67e+00&4.53e+00$\pm$2.51e+00&\cellcolor[rgb]{0.74,0.74,0.74}6.76e-01$\pm$1.47e-01\\
	\bottomrule
	\end{tabular}
	\label{tab:compare}
\end{table*}

% that's all folks
\end{document}